\documentclass[lettersize,journal]{IEEEtran}
\usepackage{amsmath,amsfonts}
\usepackage{algorithmic}
\usepackage{algorithm}
\usepackage{array}
\usepackage[caption=false,font=normalsize,labelfont=sf,textfont=sf]{subfig}
\usepackage{textcomp}
\usepackage{stfloats}
\usepackage{url}
\usepackage{verbatim}
\usepackage{graphicx}
\usepackage{cite}
\usepackage{times}
\usepackage{latexsym}

\usepackage{microtype}

\usepackage{amssymb}

\usepackage{amsthm}

\usepackage{amsmath}
\usepackage{algorithmic}
\usepackage{algorithm}

\usepackage{graphicx}
\usepackage{color}

\usepackage{multirow}

\begin{document}

\title{Adaptive Prompt Learning with Distilled Connective Knowledge for Implicit Discourse Relation Recognition}

\author{Bang Wang, Zhenglin Wang, Wei Xiang and Yijun Mo
	\IEEEcompsocitemizethanks{
		\IEEEcompsocthanksitem Bang Wang, Zhenglin Wang and Wei Xiang are with the School of Electronic Information and Communications, Huazhong University of Science and Technology (HUST), Wuhan 430074, China. Email: \{wangbang, wangzhenglin, xiangwei\}@hust.edu.cn.
		\IEEEcompsocthanksitem Yijun Mo is with the School of Computer Science and Technology, Huazhong University of Science and Technology (HUST), Wuhan 430074, China. Email: moyj@hust.edu.cn.
		\IEEEcompsocthanksitem This work is supported in part by National Natural Science Foundation of China (Grant No: 62172167). The corresponding author is Yijun Mo.
		
	}
}

\maketitle

\begin{abstract}
Implicit discourse relation recognition (IDRR) aims at recognizing the discourse relation between two text segments without an explicit connective. Recently, the prompt learning has just been applied to the IDRR task with great performance improvements over various neural network-based approaches. However, the discrete nature of the state-art-of-art prompting approach requires manual design of templates and answers, a big hurdle for its practical applications. In this paper, we propose a continuous version of prompt learning together with connective knowledge distillation, called AdaptPrompt, to reduce manual design efforts via continuous prompting while further improving performance via knowledge transfer. In particular, we design and train a few \textit{virtual tokens} to form continuous templates and automatically select the most suitable one by gradient search in the embedding space. We also design an answer-relation mapping rule to generate a few \textit{virtual answers} as the answer space. Furthermore, we notice the importance of annotated connectives in the training dataset and design a teacher-student architecture for knowledge transfer. Experiments on the up-to-date PDTB Corpus V3.0 validate our design objectives in terms of the better relation recognition performance over the state-of-the-art competitors.
\end{abstract}


%
%

\section{Introduction}
\label{Sec:Introduction}


%
%
%
%
\IEEEPARstart{I}{mplicit} Discourse Relation Recognition (IDRR) is a task to recognize the discourse relationships between a pair of text spans (also called arguments) without explicit connectives~\cite{xiang2022survey}. It is of great importance to many downstream tasks of Natural Language Processing (NLP), such as question answering~\cite{liakata2013discourse}, machine translation~\cite{guzman2014using}, information extraction~\cite{xiang2019survey}, sentiment analysis~\cite{wang2020end} and etc. Nevertheless, the IDRR task is still rather challenging task due to the absence of explicit connective. Most existing methods dealing with IDRR task are built upon the \textit{pre-train and fine-tuning} paradigm, which utilizes a well-designed Pre-train Language Model (PLM) to encode argument pairs into informative embeddings via fine-tune on the training set. But there exists some inconsistency between task-oriented objective functions of this pre-train and fine-tuning paradigm and the pre-train tasks of PLMs, which may lead to under utilization of the PLM knowledge.

\par
Recently, the \textit{pre-train, prompt, and predict} paradigm is proposed to alleviate the gap between downstream tasks and pre-train tasks of PLMs by, e.g., reformulating downstream tasks into cloze tasks. The Pre-train, prompt, and predict paradigm (also called prompt learning) has been recently investigated to tackle various NLP tasks \cite{qin2021learning, cui2021template, khashabi2020unifiedqa}. Xiang et al.~\cite{xiang2022connprompt} first adopt the prompting learning and propose the \textit{ConnPrompt} for the IDRR task with significant performance improvements over conventional neural network-based approaches. In order to better elicit a PLM knowledge, the ConnPrompt transforms the IDRR task into a cloze-task by converting an input argument pair into a natural sentence with a connective slot to be filled by the PLM. However, there are some limitations of such prompt learning-based methods.

\par
In prompt learning, prompt templates are used to transform the original input into a prompt sentence with a blank to be filled. The ability of eliciting knowledge from PLMs is determined by the design of prompt templates, while it has been shown that even a small change of templates may have great influence on the performance of prompt learning-based methods \cite{jiang2020can, gao2021making}. Traditional prompt learning employs discrete prompt templates that are likely to be suboptimal while also consuming lots of time and human efforts. Because prompt templates are to reformulate a downstream task into a cloze task, there is no need to limit templates to human-interpretable natural language. In this paper, we explore how to employ the \textit{continuous prompt template} (also called \textit{soft template}) \cite{lester2021power, zhong2021factual}, which can be fine-tuned by back-propagation to automatically search for the optimal templates for the IDRR task.

\par
In prompt learning, answer engineering aims at choosing some words to construct an answer space and designing a mapping rule from the answer space to the original output space of a specific task. However, inappropriate design of answers space and mapping rule would introduce difficulties in selecting appropriate answer words to fill in the blank of input prompt sentence, often leading to performance degradation~\cite{gao2021making}. Most existing prompt learning works construct an answer space with  manually chosen words from the English vocabulary. We call such selected words as "\textit{substantive words}". There are also no discussions on the design of an appropriate answer space for the IDRR task. In order to deal with these problems, we design an answer-relation mapping rule to automatically generate so-called "\textit{virtual answer words}", those words that do not exist in real world English vocabulary, to construct an answer space based on our pre-defined mapping rule.

\par
Additionally, it is known that the existence of an explicit connective is of great importance for the \textit{explicit discourse relation recognition} (EDRR) task \cite{pitler2008easily}, as it can provide lots of useful semantic knowledge. Fortunately, there are some manually annotated implicit connectives in the PDTB corpus, which also contain useful knowledge for discourse analysis. But as such annotated implicit connectives are not available for test and practice, they have been ignored by most existing methods for the IDRR task. We attribute this to difficult utilization of implicit connectives during model training. In this paper, we propose to implement \textit{Knowledge Distillation} (KD) via a teach-student architecture to transfer knowledge of implicit connectives in the training phase to the testing phase.

\par
In this paper, we propose a continuous version of prompt learning together with connective knowledge distillation for the IDRR task, called AdaptPrompt. Specifically, we use tunable vectors to generate continuous prompt templates that can be fine-tuned to search an optimal template by back-propagation during the training process. In addition, we design an answer-relation mapping rule to produce virtual answer words, which requires less human efforts than that of a substantive answer space. In order to extract connective knowledge from different aspects, we use both response-based and feature-based knowledge distillation techniques to convert implicit connective knowledge into soft labels and feature embeddings and transfer these knowledge from the teach model into the student model through our designed loss functions. To the best of our knowledge, this is the first paper to combine knowledge distillation with prompt learning paradigm. On the up-to-date PDTB Corpus V3.0, we conduct experiments using three advanced masked language models (BERT, DeBERTa, RoBERTa). Results show that the proposed AdaptPrompt outperforms the competing state-of-the-art models.


\par
The rest of this paper is organized as follows: Section \ref{Sec:RelatedWork} reviews related work on the IDRR task, prompt learning and knowledge distillation. Our AdaptPrompt is detailed in Section \ref{Sec:Method}. Section \ref{Sec:Experiment settings} and Section \ref{Sec:Results and Analysis} present experiment settings and results, respectively. Section \ref{Sec:Conclusion} concludes this paper with some discussions.

%
%

\section{Related Work}\label{Sec:RelatedWork}

\subsection{Implicit Discourse Relation Recognition}
Most recent solutions to the IDRR task are based on diverse neural networks~\cite{qin2017adversarial, ji2015one}. Qin et al.~\cite{qin2017adversarial} utilize two convolutional neural networks and a discriminator to construct an adversarial framework. By unearthing components' different importance to the whole input, attention mechanism has been widely adopted to extract argument representation for the IDRR task \cite{liu2016recognizing, ruan2020interactively, li2020using, liu2021importance}. For example, Liu and Li \cite{liu2016recognizing} propose the \textit{neural network with multi-level attention} (NNMA) to improve recognition of discourse relation by stabilizing some words' attention. Some researchers utilize a pre-trained language models such as BERT to learn informative argument representation \cite{ruan2020interactively, li2020using}. For example, Ruan et al. \cite{ruan2020interactively} design a cross-coupled two-channel network to construct a BERT-based propagative attention model with both self-attention and interactive-attention. Using BERT as the baseline encoder, Li et al. \cite{li2020using} propose a penalty-based loss re-estimation method to avoid misjudging unimportant words as attention-worthy words. With RoBERTa \cite{liu2019roberta}, Liu et al. \cite{liu2021importance} combine a contextualized representation module, a bilateral multi-perspective matching module and a global information fusion module to tackle implicit discourse analysis.

\subsection{Prompt Learning}
Recently prompt learning based on PLMs, such as the BERT \cite{devlin2019bert}, RoBERTa \cite{liu2019roberta}, DeBERTa \cite{he2020deberta} and etc., has been successfully applied in some NLP tasks \cite{qin2021learning, khashabi2020unifiedqa, wang2021transprompt}. For example, Qin and Eisner \cite{qin2021learning} tackle the factual probing task by designing prompt templates with some continuous vectors and fine-tuning them through gradient descent. For \textit{question and answering} task, Khashabi et al. \cite{khashabi2020unifiedqa} propose a single pre-trained QA model that transforms four different formats of QA task into a unified text generation task. Based on soft prompt template, \cite{wang2021transprompt} propose \textit{Transprompt} to capture task-agnostic and unbiased knowledge between different few-shot text classification tasks.
Recently, Xiang et al.~\cite{xiang2022connprompt} are the first to employ the prompt learning for the IDRR task with the state-of-the-art performance, where they design cloze-task hard templates with a mask for an PLM to predict substantive answer words for mapping into relation sense.

\subsection{Knowledge Distillation}
Knowledge distillation is normally used for model compression and acceleration that distills knowledge from a large and deep network (\textit{teacher model}) into a small and shallow network (\textit{student model}) \cite{hinton2015distilling}. There are in general three knowledge types for distillation, namely, response-based, feature-based, relation-based knowledge \cite{gou2021knowledge}. Response-based knowledge is the prediction after softmax layer of the teacher model, which is used to supervise the training of student model \cite{chen2017learning, meng2019conditional}. Feature-based knowledge is the output representation of both the last layer and intermediate layers. The student model imitates outputs of multiple layers of the teacher model to conduct knowledge distillation \cite{DBLP:journals/corr/RomeroBKCGB14,jin2019knowledge}. Relation-based knowledge is the relationship of different layers of the teacher model \cite{yim2017gift, zhang2018better}. Knowledge distillation has been used for some NLP tasks \cite{liu2019exploiting, yang2020model}, but to the best of our knowledge, it has not been integrated with prompt learning.

%
%

\section{The Proposed AdaptPrompt Model}\label{Sec:Method}

\subsection{Model Overview}
\begin{figure}[t]
	\centering
	\includegraphics[width=0.5\textwidth, height=0.35\textwidth]{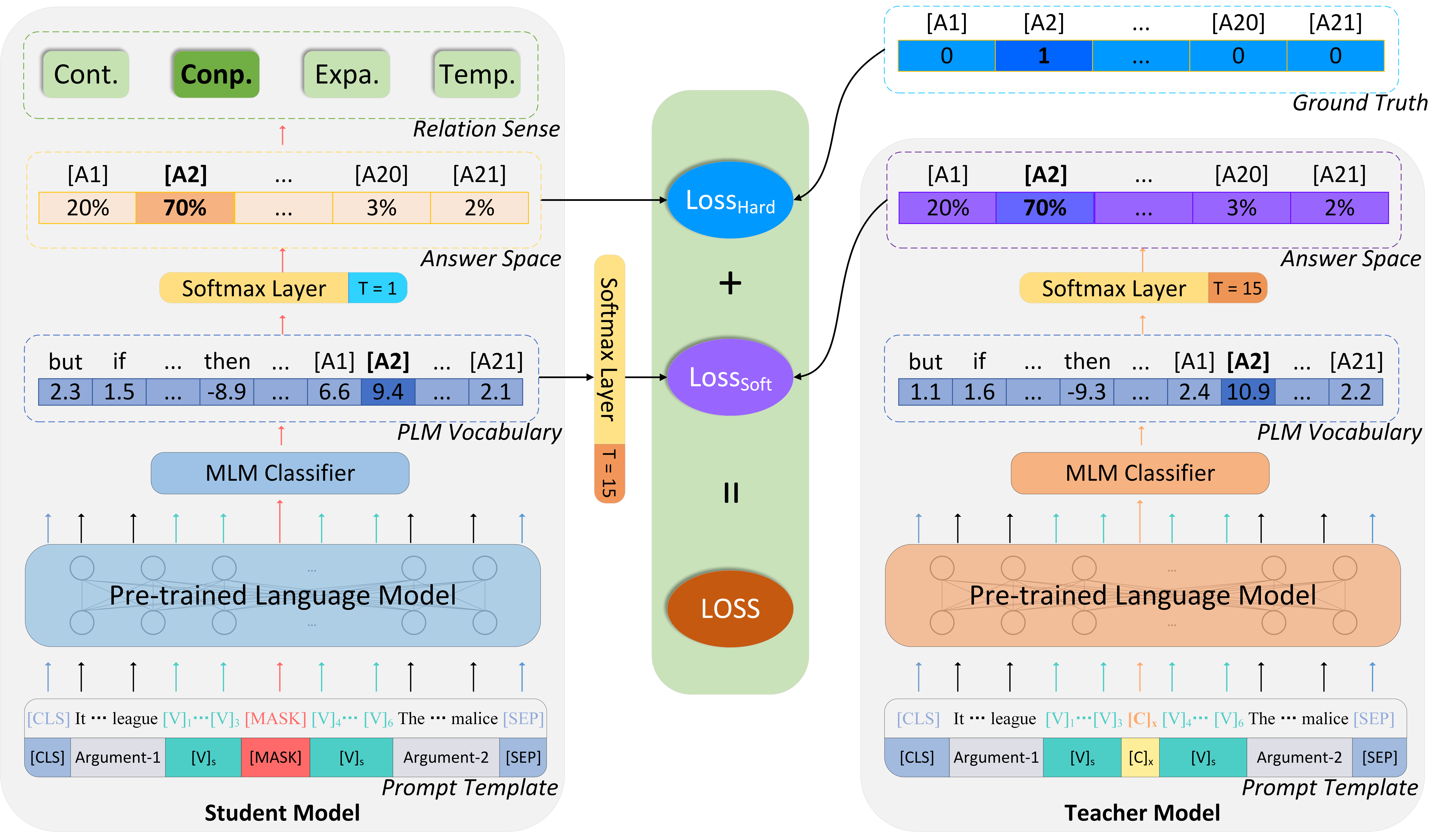}
	\caption{The overall framework of the AdaptPrompt for response-based knowledge distillation.}
	\label{Fig:response distillation}
\end{figure}

\begin{figure}[t]
	\centering
	\includegraphics[width=.5\textwidth, height=0.35\textwidth]{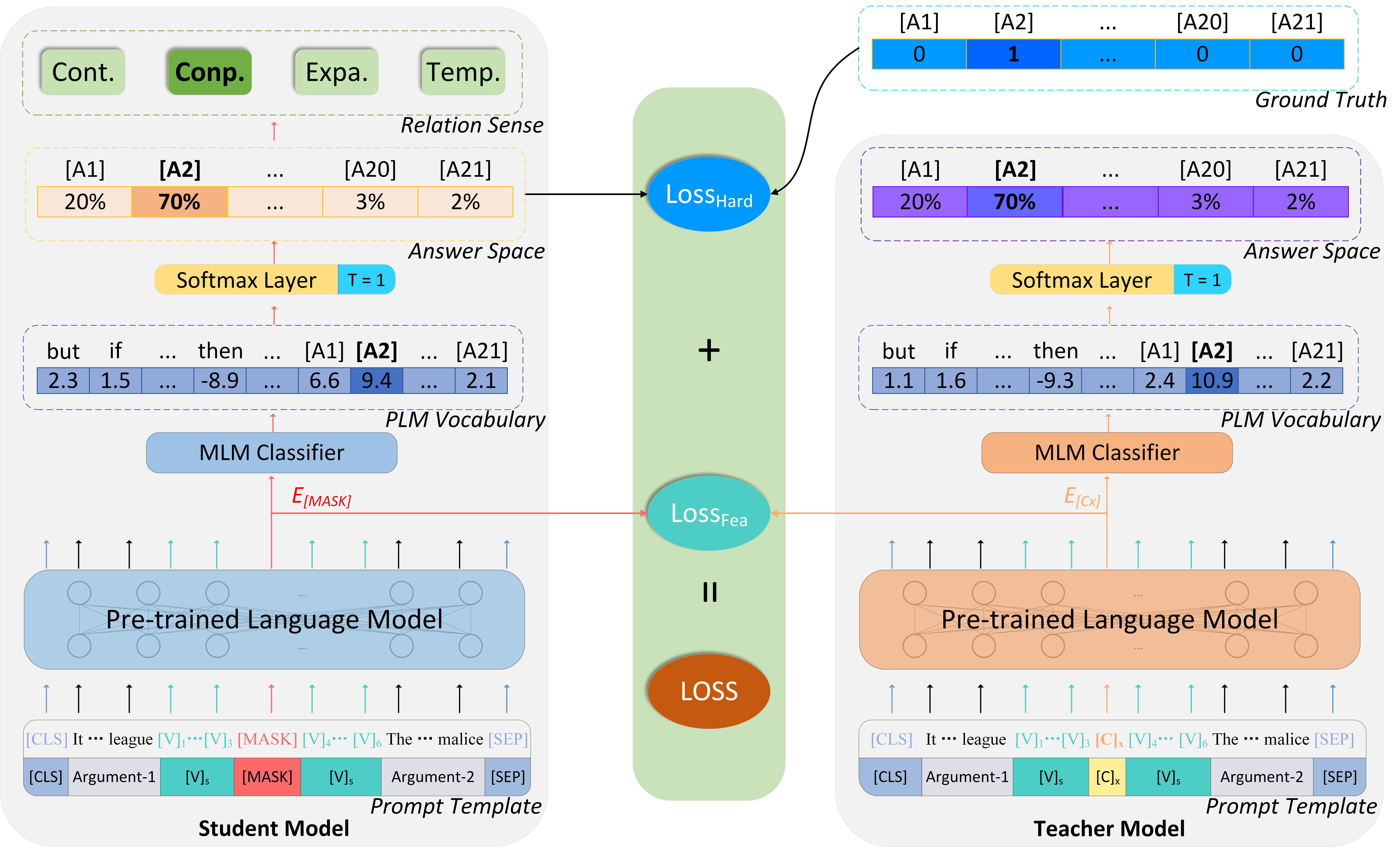}
	\caption{The overall framework of the AdaptPrompt for feature-based knowledge distillation.}
	\label{Fig:feature distillation}
\end{figure}

\begin{figure}[t]
	\centering
	\includegraphics[width=.5\textwidth, height=0.3\textwidth]{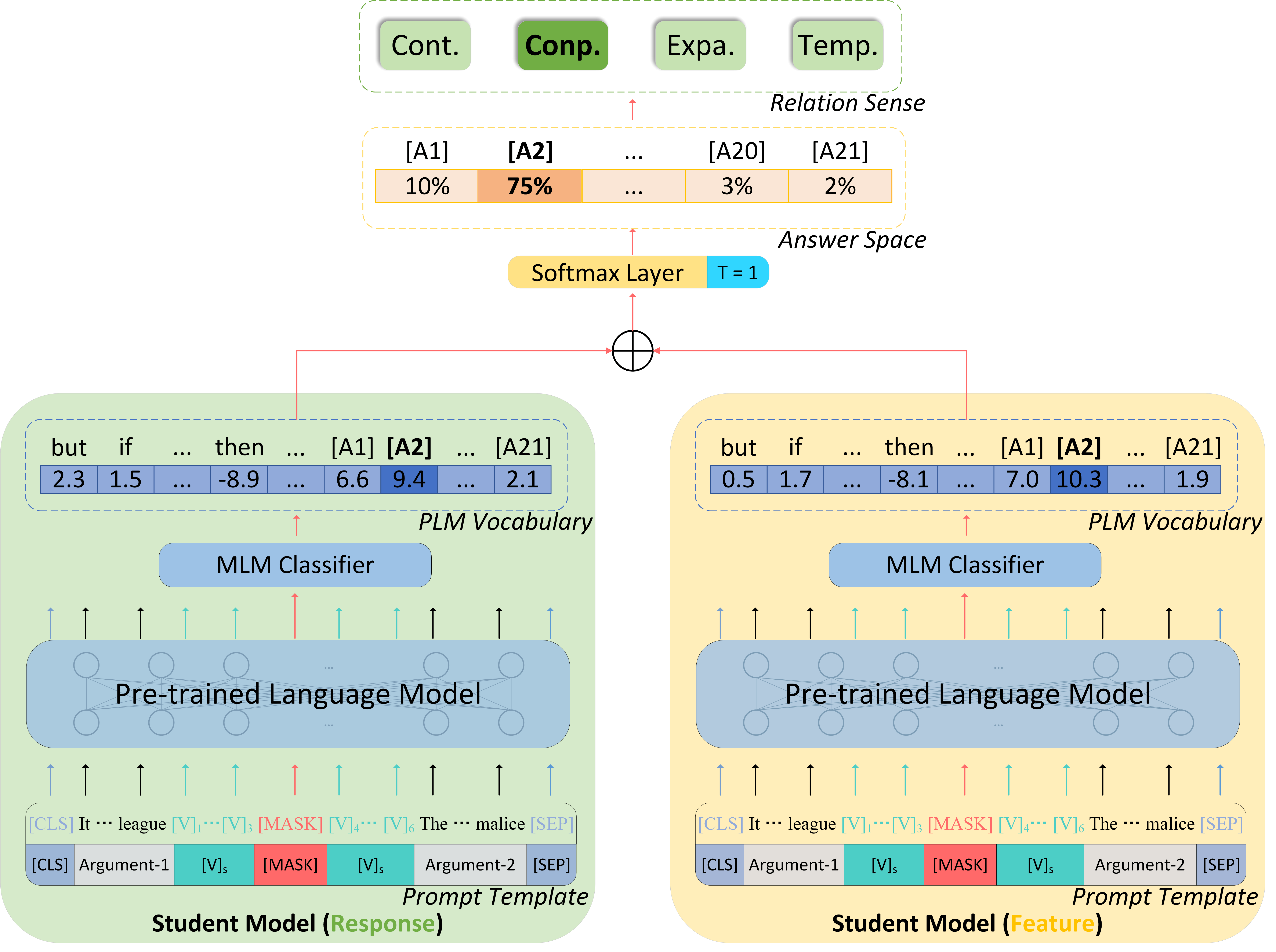}
	\caption{Illustration of fusion of two different student models.}
	\label{Fig:fusion}
\end{figure}

The proposed AdaptPrompt consists of three main components:
\begin{itemize}
	\item \textbf{Continuous Prompt Template} uses multiple learnable vectors to form a dynamic soft template which can reformulate the original argument pair into a soft prompt and automatically search for optimal templates in an embedding space. A continuous template can be fine-tuned with the whole model.
	
	\item \textbf{Virtual Answer Space} employs our answer-relation mapping rule to automatically generate a few virtual words for the answer space. These virtual words are a batch of tokens but not in the original PLM vocabulary.
	
	\item  \textbf{Connective Knowledge Distillation} extracts connective knowledge from the teacher model in the forms of prediction results after the softmax layer (response knowledge) and last hidden states of teacher's PLM (feature knowledge). With the help of ground truth, response knowledge and feature knowledge train a student model, as shown in Fig. \ref{Fig:response distillation} and Fig. \ref{Fig:feature distillation}, respectively,  to transfer connective knowledge from teacher model to student model. The final predictions are from the fusion of the two different student model, as shown in Fig. \ref{Fig:fusion}.
\end{itemize}


\subsection{Continuous Template Exploration}
Based on the intuition that the connective of two arguments can effectively indicate the relation sense of the argument pair, we design a kind of prompt template with a special token "[MASK]" representing the connective that need to be predicted by a pre-trained language model. In addition, for the purpose of eliminating the instability of manual designed discrete templates, we construct our continuous prompt template using a few learnable vectors to search for the optimal prompt template in the embedding space during the training process. As such, our continuous prompt template is composed of two arguments, multiple tunable vectors ($\mathrm{[V]_1,[V]_2,[V]_3,...}$) and some special tokens (eg. "[MASK]", "[CLS]", "[SEP]"), as follows:
\begin{align}\nonumber
	&T(Arg_1;Arg_2) = {\color[RGB]{135,206,250} \mathrm{[CLS]}} + Arg_1 + {\color[RGB]{255,215,0} \mathrm{[V]_1,...,[V]_{\frac{m}{2} }}} \\
	&+ {\color{red} \mathrm{[MASK]}}   + {\color[RGB]{255,215,0} \mathrm{[V]_{\frac{m}{2}+1 },...,[V]_m}}   + Arg_2 + {\color[RGB]{135,206,250}\mathrm{[SEP]}} \nonumber
\end{align}
where $\mathrm{[V]_i}\in \mathbb{R}^d $ and $d$ is the dimension of PLM's word embedding. The number of [V] vector $m$ is a hyper-parameter. In the training process, this tunable continuous template is fine-tuned with the whole model through back-propagation and gradient descent algorithms on the training data. In this way, formed prompt template can better elicit pre-training knowledge of a PLM on development and test data.

\subsection{Virtual Answer Generation}
In order to generate better answer space with fewer human efforts, our answer space construction first designs an answer-relation mapping rule which maps the answer space to four top-level relation senses. The PDTB 3.0 corpus not only contains four top-level relations but also has 20 second-level relations and 31 third-level relations\footnote{For those second-level relations that do not contain third-level relations, we treat themselves as corresponding third-level relations.}. All relations of different levels are detailed in Appendix Table \ref{Tab:All relations of different levels in PDTB 3.0}.

\par
Different from normal mapping rules from answer space to ultimate labels of classification (four top-level relations in this paper), our answer-relation mapping rule first maps the answer space to third-level relations and then maps third-level relations to four top-level relations. Our mapping rule has two main principles: (1) For most of the third-level relations, we use an answer in the answer space to represent a third-level relation. (2) For those third-level relations with too few instances and have other semantically similar relations, we use an answer in the answer space to represent the instance-few relations together with their respective similar relations. The details of our answer-relation mapping rule are presented in Appendix Table \ref{Tab:Virtual answer space construction rules}.

\par
The second step of our answer space construction is to create some virtual answers to form the answer space based on aforementioned answer-relation mapping rule. We create some new tokens ($\mathrm{[A]_1,[A]_2,...,[A]_{21}}$) according to the mapping rule and add them into the PLM's vocabulary as the answer space. Because these answers do not exist in English language of real world, we call them "virtual answers". The details are presented in Table \ref{Tab:Virtual answer space construction rules}.

\par
The last step is to initialize these virtual answers. In order to initialize their embeddings to represent third-level relations, we set the initial word embedding $E_{\mathrm{[A]_x}}$ of each virtual answer to be the average of word embeddings of each implicit connective $\mu_i$ that appear in the instances of the corresponding third-level relation:
\begin{equation}
	E_{\mathrm{[A]_x}} = \frac{\sum_{i=1}^{s_x}E_{\mu_i}}{s_x},
\end{equation}
where $s_x$ is the number of instances of the third-level relation that corresponds to $\mathrm{[A]_x}$.

\par
\begin{figure}[t]
	\centering
	\includegraphics[width=.4\textwidth, height=.3\textwidth]{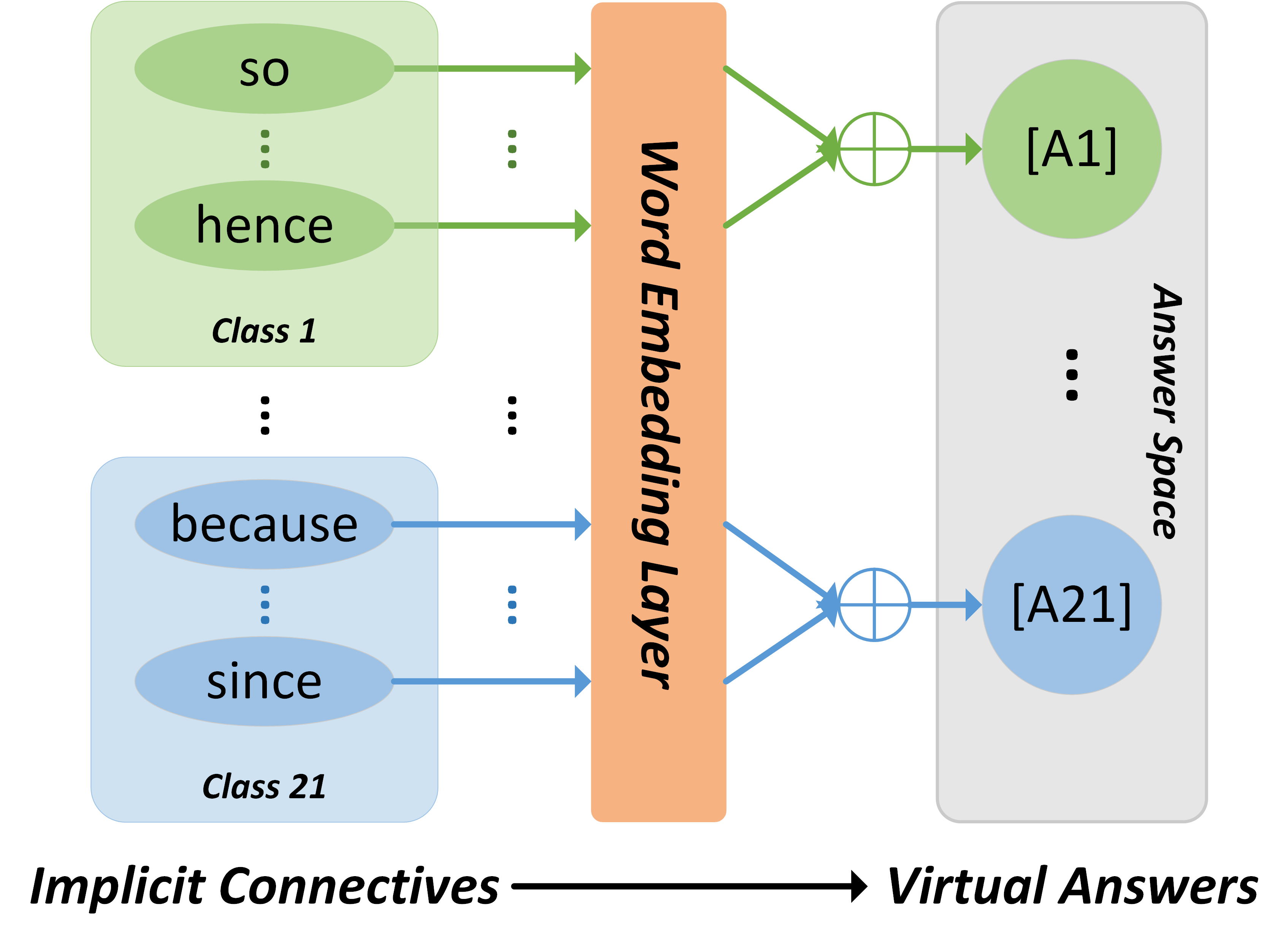}
	\caption{Illustration of virtual answer initialization algorithm}
	\label{Fig:virtual answer initialization}
\end{figure}

For example, $\mathrm{[A]_2}$ corresponds to the \textit{Comparison.Contrast}, so the initial word embedding of $\mathrm{[A]_2}$ is set to be the average of word embeddings of 742 instances' connectives in the \textit{Comparison.Contrast} relation; $\mathrm{[A]_5}$ corresponds to \textit{Contingency.Purpose.Arg2-as-goal} and \textit{Contingency.Purpose.Arg1-as-goal} so the initial word embedding of $\mathrm{[A]_5}$ is set to be the average of word embeddings of totally 1105 instances' connectives in the \textit{Contingency.Purpose.Arg2-as-goal} and \textit{Contingency.Purpose.Arg1-as-goal}. The initialization process is presented in Fig. \ref{Fig:virtual answer initialization}.

\subsection{Connective knowledge Distillation}
Teacher model uses the aforementioned answer space structure and mapping rule for prediction. In order to incorporating implicit connective information into soft labels and last hidden embeddings, the teacher model adopts continuous prompt template but replaces the "[MASK]" token with the corresponding implicit connective of every sample. Thus, the adapted prompt template is composed of two arguments, multiple tunable vectors, a specific implicit connective and some special tokens (eg. "[CLS]", "[SEP]") as follows:
\begin{align}\nonumber
	&T(Arg_1;Arg_2) = {\color[RGB]{135,206,250} \mathrm{[CLS]}}    + Arg_1  + {\color[RGB]{255,215,0} \mathrm{[V]_1,...,[V]_{\frac{m}{2} }}} \\
	&+ {\color[RGB]{50,205,50} \mathrm{[C]_x}}  + {\color[RGB]{255,215,0} \mathrm{[V]_{\frac{m}{2}+1 },...,[V]_m}}   + Arg_2 + {\color[RGB]{135,206,250}\mathrm{[SEP]}} \nonumber
\end{align}

\par
\begin{figure}[t!]
	\centering
	\includegraphics[width=.4\textwidth, height=.3\textwidth]{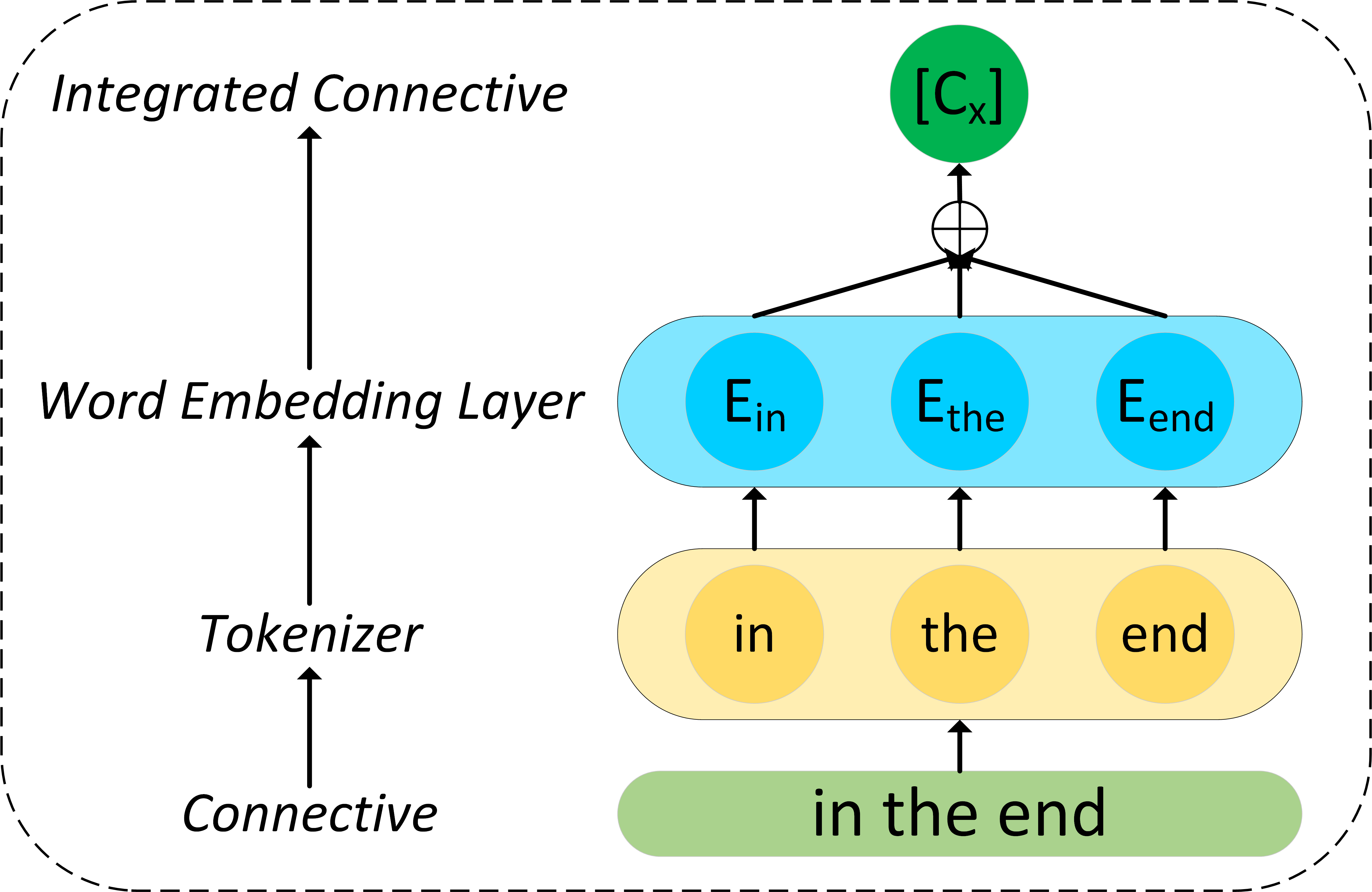}
	\caption{Illustration of integrated connective initialization}
	\label{Fig:integrated connective}
\end{figure}

Different from "[MASK]" token treated as just one token, most implicit connectives in the PDTB corpus contain more than one token after tokenization (e.g., the connective \textit{"on the contrary"} is tokenized into three tokens for the BERT-base: \{\textit{"on", "the", "contrary"}\}). For treating each connective as one token to make the form of teacher model's prompt similar to student model, we create some new tokens, called "\textit{integrated connective}", for each implicit connective in the overall vocabulary. As shown in Fig. \ref{Fig:integrated connective}, we use these integrated connectives ($\mathrm{[C]_1, [C]_2, ...}$) to represent original connectives and initialize every integrated connective token $E_{\mathrm{[C]_x}}$ with the average of all tokens' word embedding of the original connective:
\begin{equation}
	E_{\mathrm{[C]_x}} = \frac{\sum_{i=1}^{t_x}E_{\lambda _i}}{t_x}
\end{equation}
where $E_{\lambda _i}$ is the word embedding of $i$-th token $\lambda _i$ of an original connective and $t_x$ denotes the number of all tokens of the original connective.

\par
After transferring the original argument pair into continuous prompt, the PLM of teacher model estimates scores of each token in the whole PLM's vocabulary for the integrated connective token "$\mathrm{[C]_x}$". The soft labels (probability distribution) are computed by applying the softmax function with a pre-defined temperature $T$ into all answers' scores:
\begin{equation}
	P(a_i,T) = \frac{e^{a_i/T}} {\sum_{j=1}^{n}e^{a_j/T}},
\end{equation}
where $a_i$ denotes the output score of the $i$-th answer, $n$ the answer space size and $P(a_i,T)$ the output probability of the $i$-th answer with the distillation temperature $T$. The feature knowledge is the last hidden embeddings of integrated connective tokens which are generated by the PLM's last hidden layer.

\par
Two student models with the same continuous template, virtual answer space and network architecture utilize soft labels and hidden embeddings, respectively, which include abundant implicit connective information for model training with assistance of the ground-truth labels. According to the results on the development set, the best response-based student model and feature-based student model are chosen to form the final prediction results. The predict logits on the test set of two different student models are fused by average pooling and softmax function to generate the final prediction label, so as to enjoy the advantages of two types of knowledge distillation.
\par

\subsection{Model Training}
\textbf{Student Model} In the training procedure, we use the back-propagation algorithm to tune all parameters of the whole student model on the IDRR training dataset.

\par
For \textit{response-based student model}, there are two kinds of predicted labels: One is computed by the normal softmax function denoted, as $\hat{y}^{(1)}$; The other is computed by the adapted softmax function with the same temperature as that of the teacher model, denoted as $\hat{y}^{(T)}$. The overall loss function is composed of both the hard loss and soft loss:
\begin{equation}
	Loss_{student}^{response}=\alpha Loss_{hard}+(1-\alpha )Loss_{soft},
\end{equation}
where $\alpha$ ($0<\alpha<1$) is the balance coefficient. For the hard loss function, we adopt the cross entropy loss and take the predicted label (with $T=1$) and gold label as input:
\begin{equation}
	Loss_{hard}=-\frac{1}{K}\sum_{k=1}^{K} y_k\log{\hat{y}_{k}^{(1)}},
\end{equation}
where $y_k$ and $\hat{y}_{k}^{(1)}$ are the gold label and predicted label by the student model with the normal softmax of $k$-th training instance, respectively. For the soft loss function, we apply the Kullback-Leibler (KL) divergence loss and take the predicted label (with $T=t$) and soft label (of teacher model) as input:
\begin{equation}
	Loss_{soft}=\frac{1}{K}\sum_{k=1}^{K} \bar{y}_k\log{\frac{\bar{y}_k}{\hat{y}_{k}^{(T)}}},
\end{equation}
where $\bar{y}_k$ and $\hat{y}_{k}^{(T)}$ are the soft label and predicted label with the adapted softmax with $T$ of $k$-th training instance, respectively.

\par
For the \textit{feature-based student model}, the overall loss function contains two parts: hard loss (the same as response-based student model) and the feature loss:
\begin{equation}
	Loss_{student}^{feature}= Loss_{hard} + \beta Loss_{feature},
\end{equation}
where $\beta$ is a weight coefficient to balance the importance of hard loss and feature loss. For the feature loss function, we adopt the Mean Square Error (MSE) loss and take the last hidden embeddings of the student model and teacher model as input:
\begin{equation}
	Loss_{feature}=MSE(E_{[MASK]}, E_{[C]_{x}}),
\end{equation}
where $E_{[MASK]}$ is the last hidden embeddings of [MASK] token of student model and $E_{[C]_{x}}$ is the last hidden embeddings of $[C]_x$ token of teacher model. We use the AdamW optimizer \cite{loshchilov2018decoupled} with $L2$ regularization for model training.

\par
\textbf{Teacher Model} In the training stage, we use the cross entropy loss as the cost function and train teacher model to predict labels close to gold labels. For each instance in the training set, we save the distilled probability distribution of all virtual answers as soft labels and last hidden embeddings of $[C]_x$ token as feature knowledge, when teacher model attains the best performance on the development set.

%
%
\section{Experiment Settings}\label{Sec:Experiment settings}

\subsection{The PDTB 3.0 Dataset}
We conduct our experiment on the latest version 3.0 of the Penn Discourse TreeBank (PDTB) corpus, which was developed by the University of Pennsylvania on March 2019 and updated on February 2020 through the Linguistic Data Consortium (LDC). Following Jin's data partition in PDTB 2.0 \cite{ji2015one}, we treat sections 0-1 as the development set, sections 2-20 as the training set and sections 21-22 as the testing set. Our experiments focus on the four top-level relation classification, including \textit{Comparison, Contingency, Expansion, Temporal}. Table \ref{Tab:PDTB dataset} presents the statistics of implicit discourse relation instances in the PDTB 3.0 corpus.

\begin{table}[t]
	\centering
	\caption{Statistics of instances with four top-level implicit discourse relations in PDTB 3.0 corpus.}
	\begin{tabular}{ c|cccc|c}
		\hline
		Relation  & Expa. & Comp. & Cont. & Temp. & Total\\
		\hline
		Train  & 8645 & 1937 & 5916 & 1447 & 17945 \\
		Dev.  & 748 & 190 & 579 & 136 & 1653 \\
		Test & 643 & 154 & 529 & 148 & 1474 \\
		\hline
	\end{tabular}
	\label{Tab:PDTB dataset}
\end{table}

\subsection{Pre-trained Language Models}
We conduct our experiments on three pre-trained masked language models for comparison:
\begin{itemize}
	\item \textbf{BERT} \cite{devlin2019bert}: The most representative PLM proposed by google, whose pre-training tasks are composed of a \textit{masked word prediction} task and a \textit{next sentence prediction} task.
	\item \textbf{RoBERTa} \cite{liu2019roberta}: A modified BERT PLM proposed by Facebook, which is pre-trained on larger dataset without the \textit{next sentence prediction} task.
	\item \textbf{DeBERTa} \cite{he2020deberta}: The masked PLM proposed by Microsoft, which use a disentangled attention mechanism, an enhanced mask decoder and virtual adversarial training strategy for the model pre-training.
\end{itemize}

\subsection{Comparison Models}
We compare our AdaptPrompt with the following advanced models:
\begin{itemize}
	\item \textbf{DAGRN} \cite{chen2016implicit} extracts semantic interaction between word pairs using a gated relevance network.
	\item \textbf{NNMA} \cite{liu2016recognizing} combines the attention mechanism and external memories to imitate reading the arguments repeatedly.
	\item \textbf{IPAL} \cite{ruan2020interactively} uses a cross-coupled two-channel network to make use of self-attention and interactive-attention mechanisms.
	\item \textbf{PLR} \cite{li2020using} uses penalty coefficients in the computation of loss to regulate the attention learning.
	\item \textbf{BMGF} \cite{liu2021importance} combines contextualized representation, bilateral multi-perspective matching and global information fusion to tackle implicit discourse analysis.
	\item \textbf{MANF} \cite{xiang2022encoding} utilizes dual attention network to fuse two kinds of attentive representation as semantic connection and generates linguistic evidence by an offset matrix network.
	\item \textbf{ConnPrompt} \cite{xiang2022connprompt} adopts prompt learning paradigm for the IDRR task and designs three kinds of discrete prompt templates and a static answer space.
\end{itemize}

\subsection{Parameter Settings}
Table \ref{Tab:PLM config} provides the configurations of the three PLMs. We use the PyTorch framework and \textit{transformers} package of HuggingFace to build our model, including all pre-trained language models. We run experiments on NVIDIA GTX TITAN X GPUs with CUDA 10.1. We set the length of each input prompt to 100 tokens and the maximum length of argument-1 to 50 tokens, which exceeds most arguments' length in the PDTB 3.0. We use mini-batch method for training and set batch size to 32. We also use four learning rates, including 5e-5, 2e-5, 1e-5, 5e-6, to search for the best model performance. In the main results, the distillation temperature $T$ is set to 10, 15, 20; the balance coefficient $\alpha$ in the loss function of response-based student model is set to 0.3, 0.5, 0.7; and the weight coefficient $\beta$ in loss function of feature-based student model is set to 5e-4, 2e-4, 1e-4, 5e-5. The number of [V] tokens in prompt template is set to 6, and 3 [V]s before "[MASK]" and 3 [V]s behind "[MASK]", respectively.

\begin{table}[t]
	\centering
	\caption{The configuration of three pre-trained masked language models.}
	\begin{tabular}{ c|cccc}
		\hline
		PLM  & Model & Vocab. size & Layer & Dim.\\
		\hline
		BERT  & bert-base-uncased &30522 & 12 & 768 \\
		RoBERTa  & roberta-base &50265 & 12 & 768 \\
		DeBERTa & debert-base  &50265 & 12 & 768 \\
		\hline
	\end{tabular}
	\label{Tab:PLM config}
\end{table}

%
%
\section{Result and Analysis}\label{Sec:Results and Analysis}
\subsection{Overall Result}
We use the two standard performance metrics for the IDRR task: \textit{F1} score and Accuracy (ACC). Table \ref{Tab:Overall Results} compares the overall results of our AdaptPrompt and the competitors. Specially, we use the best results of the ConnPrompt model on its single-prompt setting for fair comparison. The first three competitors, viz., DAGRN, NNMA and MANF-Word2Vec, make use of Word2Vec and Glove language model to transfer English words into static word embeddings. Dynamic and contextual word embeddings, provided by some Transformer-based pre-trained language models (BERT, RoBERTa, DeBERTa), are used by other listed models.

\par
The IPAL, PLR and BMGF outperform the DAGRN and NNMA; While the MANF-BERT model outperforms MANF-Word2Vec model. This can be attributed to their utilization of Transformer-based PLMs as well as more advanced network architecture. The ConnPrompt model outperforms the IPAL, PLR and MANF-BERT model with significant improvements, and it has obvious advantages even using the same PLM as other models. These results can be attributed to its employment of the prompt learning paradigm that makes better use of PLM knowledge than the traditional fine-tuning paradigm.

\par
Our AdaptPrompt attains greater performance improvements over the ConnPrompt on the three PLMs, and our AdaptPrompt with RoBERTa outperforms all the other models. We attribute the improvements to its employment of continuous prompt template, virtual answer space and connective knowledge distillation. Continuous prompt templates allow us to search for a better template than manual designed template. Virtual answer space automatically generates virtual answer words which have strong connection with the corresponding relation senses than those of manually-chosen substantive answer words. Connective knowledge distillation densely utilizes the implicit connective information of each training instance as response knowledge and feature knowledge to further improve the performance of AdaptPrompt.


\begin{table}[t]
	\centering
	\caption{Overall results of different models for IDRR on PDTB 3.0 corpus. The boldface is the best among all results, the underline is the second best results and * denotes the best results among all models of a specific PLM (BERT, DeBERTa or RoBERTa).}
	\begin{tabular}{ c|c|c|c}
		\hline
		PLM & Model & Acc (\%) & F1 (\%) \\
		\hline
		\multirow{2}{*}{Word2Vec} & DAGRN (ACL, 2016) & 57.33 & 45.11 \\
		& MANF (ACL, 2022) & 60.45 & 53.14 \\
		\hline
		Glove & NNMA (EMNLP, 2016) & 57.67 & 46.13 \\
		\hline
		\multirow{5}{*}{BERT} & IPAL (COLING, 2020) & 57.33 & 51.69 \\
		& PLR (COLING, 2020) & 63.84 & 55.74 \\
		& MANF (ACL, 2022) & 64.04 & 56.63 \\
		& ConnPrompt (COLING, 2022) & 69.74 & 63.95 \\
		& \textbf{Our Model} & 70.28* & 66.16* \\
		\hline
		\multirow{2}{*}{DeBERTa} & ConnPrompt (COLING, 2022) & 72.66 & 67.98 \\
		& \textbf{Our Model} & \underline{74.56*} & \underline{70.67*} \\
		\hline
		\multirow{3}{*}{RoBERTa} & BMGF (IJCAI, 2020) & 69.95 & 62.31 \\
		& ConnPrompt (COLING, 2022) & 74.36 & 69.91 \\
		& \textbf{Our Model} & \textbf{76.12*} & \textbf{71.79*} \\
		\hline
	\end{tabular}
	\label{Tab:Overall Results}
\end{table}

\subsection{Ablation Study}
Our AdaptPrompt has three main modules: \textit{Connective Knowledge Distillation} (CKD), \textit{Continuous Prompt Template} (CPT) and \textit{Virtual Answer Space} (VAS). We conduct our ablation experiments according to the three modules. Table \ref{Tab:Ablation Results} presents the ablation results for different versions of the proposed AdaptPrompt on three PLMs.

\par
The AdaptPrompt-RD model removes the response-based knowledge distillation and treats feature-based student model as the final prediction model. Similarly, the AdaptPrompt-FD model removes the response-based knowledge distillation. As for the AdaptPrompt-RD-FD model, we train the student model without the supervision of teacher model and connective knowledge but only hard labels.

\par
The AdaptPrompt-CPT model utilizes discrete prompt template as ConnPrompt \cite{xiang2022connprompt} instead of continuous prompt templates, while the AdaptPrompt-VAS model employs substantive answer space the same as ConnPrompt rather than virtual answer space in this paper. The rest four versions of AdaptPrompt are combination of the above modules for ablation (for example, AdaptPrompt-CPT-RD model has no continuous prompt template and response-based student model).

\par
Compared with model AdaptPrompt-RD-FD without supervision of connective knowledge, the models AdaptPrompt-RD and AdaptPrompt-FD have higher performance on IDRR task. The results suggest that connective knowledge is significant for IDRR task and knowledge distillation is a suitable way of transferring connective knowledge, although there are different types of knowledge. In addition, we also notice that AdaptPrompt model (complete model) outperforms any model with single type of connective knowledge. We attribute this to advantages of our ensemble strategy for connective knowledge distillation. The AdaptPrompt model with ensemble of two types of knowledge have more thorough information of implicit connective from different aspects, which is important for IDRR.

\par
The AdaptPrompt model (complete model) outperforms AdaptPrompt-CPT model. The results show that continuous prompt template can automatically search for a better prompt template by gradient descent than manually designed templates that is used by AdaptPrompt-CPT model. Also, AdaptPrompt-CPT model has higher performance than AdaptPrompt-CPT-RD and AdaptPrompt-CPT-FD model, which means even in situation of discrete prompt template, ensemble of connective knowledge from different aspects still benefit to model training.

\par
The AdaptPrompt model (complete model) outperforms AdaptPrompt-VAS model. It can be attributed to the utilization of virtual answer space. Created by average of multiple connectives, virtual words for answer space are more representative than substantive answers chosen from real world connectives. Appropriate answer space have great positive impact on prompt-based models. Moreover, AdaptPrompt-VAS model attains better performance than AdaptPrompt-VAS-RD and AdaptPrompt-VAS-FD model. It can be attributed that ensemble strategy of knowledge distillation still works even on substantive answer space setting.

\begin{table*}[ht!]
	\centering
	\caption{The results of ablation study of our model on three PLMs. CPT is the abbreviation of Continuous Prompt Template; VAS is the abbreviation of Virtual Answer Space; CKD is the abbreviation of Connective Knowledge Distillation; RD is the abbreviation of Response-based Distillation; FD is the abbreviation of Feature-based Distillation.}
	\begin{tabular}{ c|c|cc|cc|cc }
		\hline
		\multirow{2}{*}{Module} & \multirow{2}{*}{Model} & \multicolumn{2}{|c|}{BERT} & \multicolumn{2}{|c|}{DeBERTa} & \multicolumn{2}{|c}{RoBERTa} \\
		\cline{3-8}
		& & Acc & F1 & Acc & F1 & Acc & F1 \\
		\hline
		- & AdaptPrompt (our model) & 70.28 & 66.16 & 74.56 & 70.67 & 76.12 & 71.79 \\
		\hline
		\multirow{3}{*}{CKD} & AdaptPrompt-RD & 70.01 & 65.64 & 73.41 & 69.66 & 75.58 & 71.49 \\
		& AdaptPrompt-FD & 70.42 & 65.75 & 73.61 & 69.57 & 75.03 & 71.41 \\
		& AdaptPrompt-RD-FD & 69.20 & 64.46 & 71.64 & 68.41 & 74.97 & 70.87 \\
		\hline
		\multirow{3}{*}{CPT} & AdaptPrompt-CPT & 69.81 & 64.80 & 74.15 & 69.97 & 75.44 & 70.90 \\
		& AdaptPrompt-CPT-RD & 69.88 & 64.16 & 73.07 & 69.23 & 75.31 & 70.88 \\
		& AdaptPrompt-CPT-FD & 69.47 & 64.39 & 72.86 & 68.61 & 74.63 & 70.24 \\
		\hline
		\multirow{3}{*}{VAS} & AdaptPrompt-VAS & 70.56 & 65.41 & 74.02 & 69.94 & 75.64 & 71.50 \\
		& AdaptPrompt-VAS-RD & 69.95 & 64.30 & 73.88 & 69.31 & 74.83 & 70.86 \\
		& AdaptPrompt-VAS-FD & 69.88 & 64.13 & 72.66 & 68.81 & 74.69 & 70.55 \\
		\hline
	\end{tabular}
	\label{Tab:Ablation Results}
\end{table*}

\subsection{Effections of Knowledge Distillation Temperature}
\begin{figure}[h!]
	\centering
	\subfloat[\small{Accuracy of different temperature in knowledge distillation.}]{
		\includegraphics[scale=0.27,trim=10 0 40 35,clip]{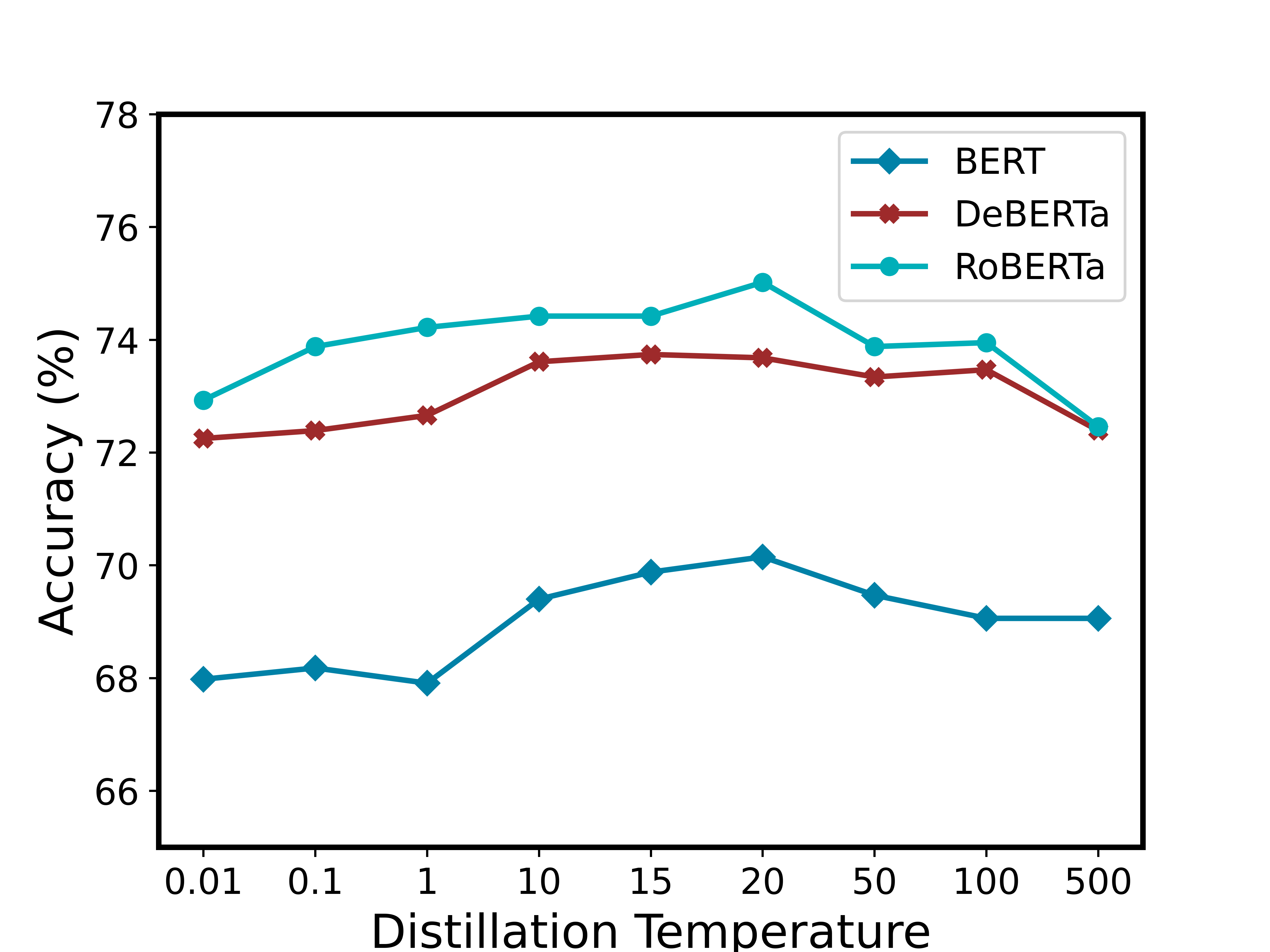}
		\label{Fig:Acc of Temperature}
	}
	\quad
	\subfloat[\small{F1-score of different temperature in knowledge distillation.}]{
		\includegraphics[scale=0.27,trim=10 0 40 35,clip]{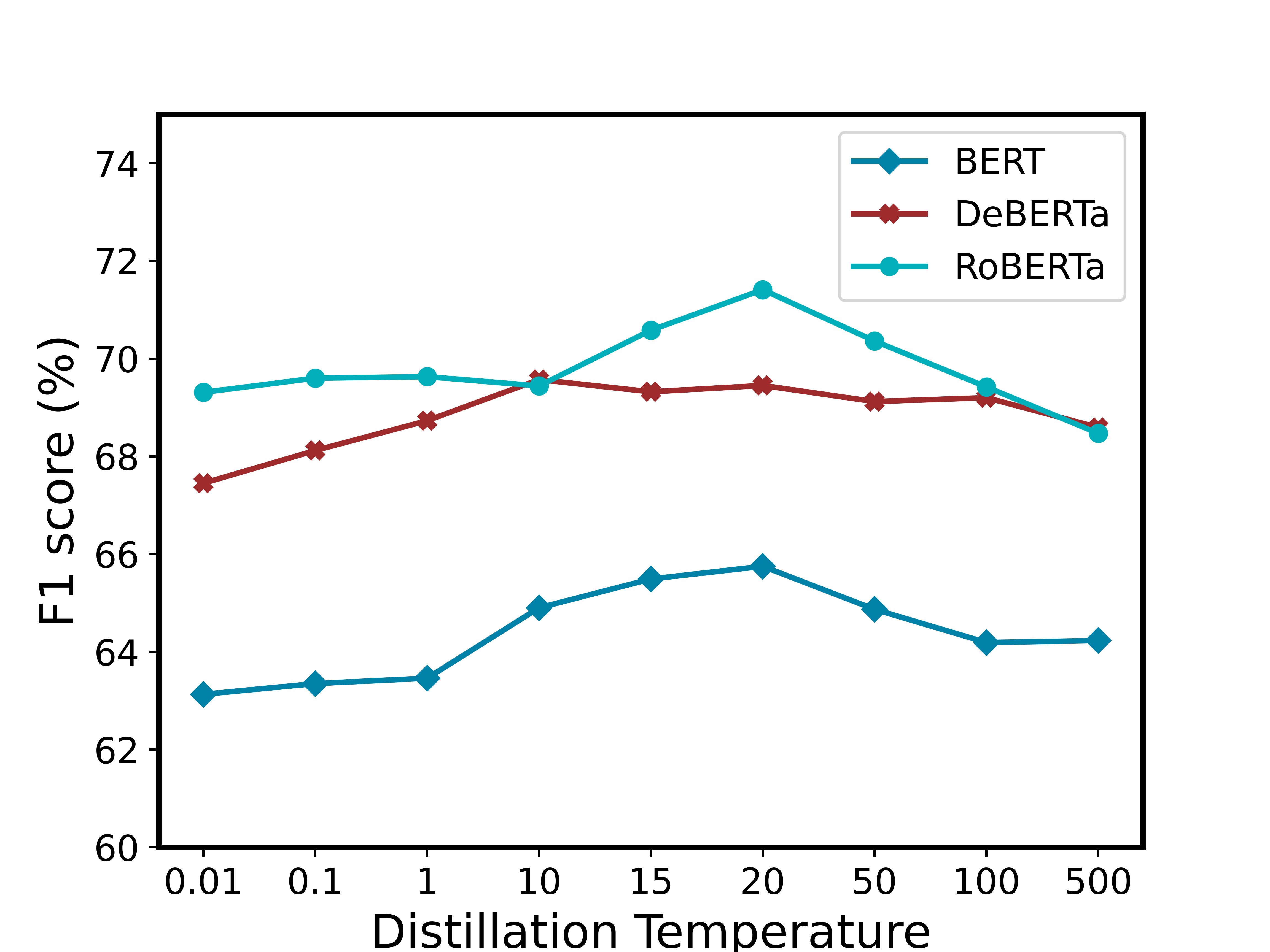}
		\label{Fig:F1 of Temperature}
	}
	\caption{The performance of different temperature of Response-based Knowledge Distillation.}
	\label{Fig:Results of Temperature}
\end{figure}

Distillation temperature $T$ is a hyper-parameter for the training of response-based student model, which controls the importance of each class in soft labels by scaling softmax function. In order to highlight effects of distillation temperature for response-based distillation, we remove the part of feature-based student model and only use response-based student model for experiments. Figs.~\ref{Fig:Acc of Temperature} and \ref{Fig:F1 of Temperature} present the Acc and F1 results of using different temperatures.


\par
We can observe that when the temperature $T$ is higher than 20, the performance of AdaptPrompt on all three PLMs decreases. This is because if the distillation temperature is too high, negative classes are treated almost as important as the positive class, which means soft labels are prone to be average among all classes. This kind of soft labels contains few useful knowledge from the teacher model. In addition, when the temperature $T$ is lower than 10, the AdaptPrompt performs worse. We attribute this to the ignorance of negative classes. If the temperature used for distillation is too low (especially lower than 1), negative classes of each soft label have no importance, which makes soft labels tend to be one-hot appearance like gold labels. If the teacher model cannot predict each instance in the training set correctly, such one-hot like soft labels contain some wrong information that may be harm to the student model training.

\subsection{Effections of Teacher Model}
\begin{figure}[h!]
	\centering
	\subfloat[\small{Accuracy of different ways of utilizing connective knowledge.}]{
		\includegraphics[scale=0.28,trim=5 0 5 5,clip]{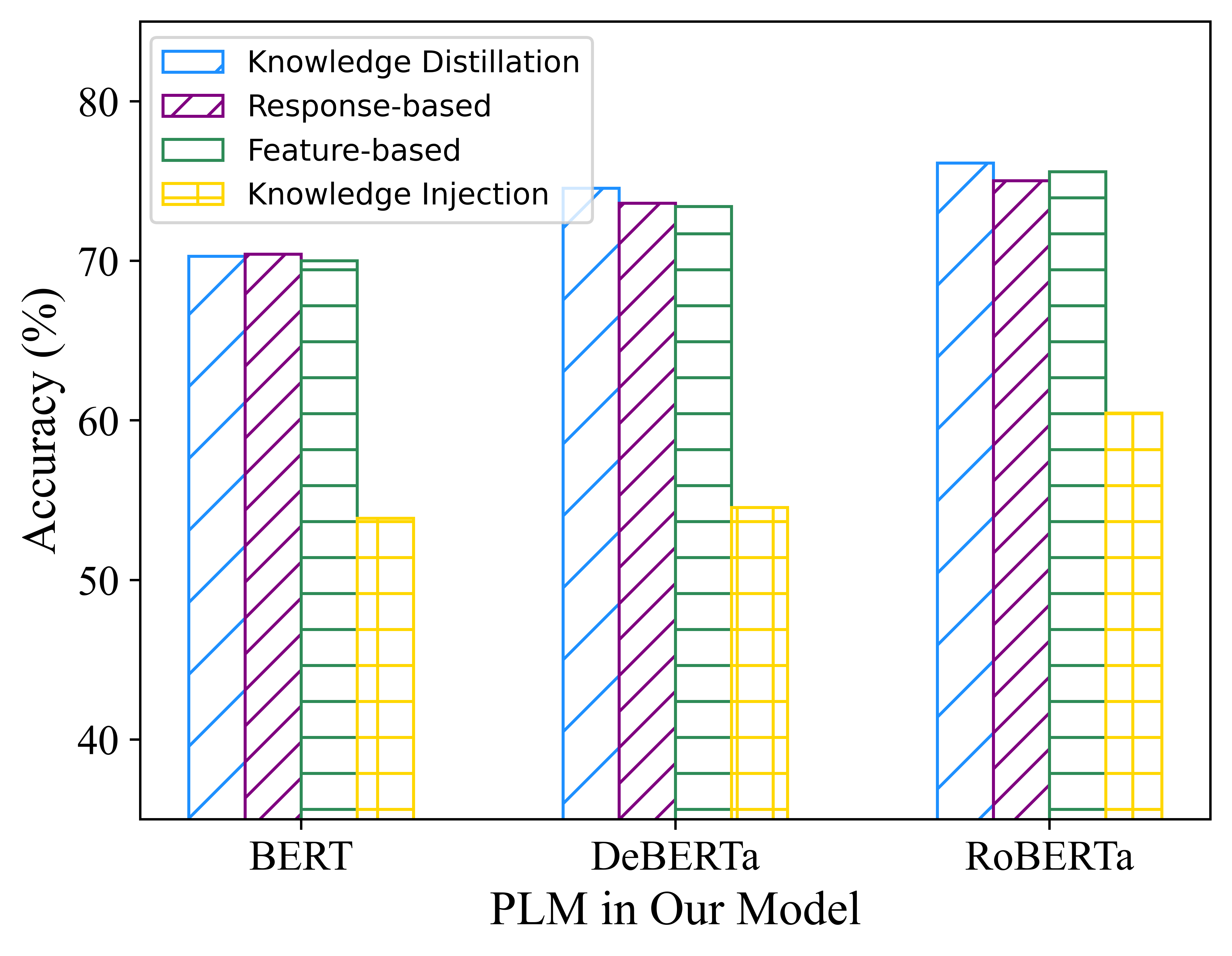}
		\label{Fig:Acc of T-S Architecture}
	}
	\quad
	\subfloat[\small{F1-score of different ways of utilizing connective knowledge.}]{
		\includegraphics[scale=0.28,trim=5 0 5 5,clip]{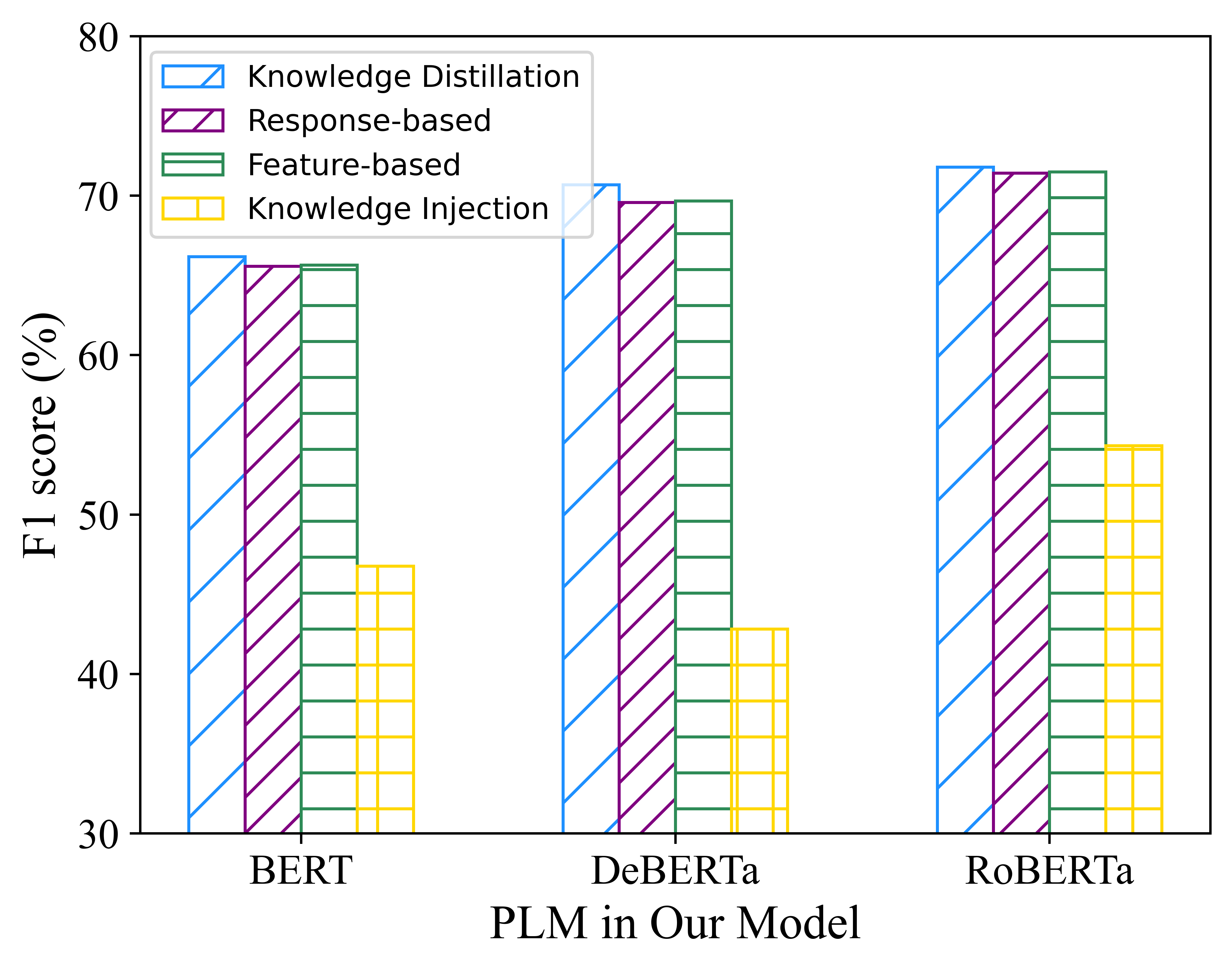}
		\label{Fig:F1 of T-S Architecture}
	}
	\caption{The performance of different ways for utilizing annotated implicit connectives.}
	\label{Fig:Results of T-S Architecture}
\end{figure}

In this section, we conduct experiments to compare two ways of utilizing knowledge of annotated connectives by different T-S architectures ("Knowledge Distillation", "Response-based" and "Feature-based" in Fig. \ref{Fig:Results of T-S Architecture}) and by directly using annotated connectives in the model training without the help of teacher model (we called this "Knowledge Injection" in Fig. \ref{Fig:Results of T-S Architecture}). "Knowledge Distillation" in Fig. \ref{Fig:Results of T-S Architecture} is our complete strategy of utilizing T-S architecture, while "Response-based" and "Feature-based" are only employ response knowledge and feature knowledge for student model respectively.

\par
The way of knowledge injection is as follows: We eliminate the part of teacher model and treat student model as the whole model. In the training process, we replace "[MASK]" token with annotated connectives in the input prompt template and make the PLM to predict answers according to embeddings of connectives. In the development and testing process, the prompt template still uses "[MASK]" token for prediction without connectives.

\par
It can be observed that utilizing the T-S architecture has better performance than directly using implicit connectives of knowledge injection by a large margin. In fact, the latter way of using annotated connectives is to transfer an IDRR task into an EDRR task during the training process, but remaining an IDRR task during the testing process. In training, the model of knowledge injection may focus on constructing the relationship between connectives and discourse senses while ignoring the interaction between two arguments to some extent, which make it hard to predict discourse senses of instances without connectives in testing. The discrepancy between the training and testing process leads to its performance degradation.

\subsection{Virtual Answer Space Design Effections}
In our AdaptPrompt, we design an answer-relation mapping rule, which assigns a virtual answer to a third-level relations in most cases and assign a virtual answer to multiple third-level relations in some cases. We can also have three other mapping rules. The first \textit{Verbalizer 1} assigns a virtual answer to each top-level relation with 4 virtual answers. The second \textit{Verbalizer 2} assigns a virtual answer to each second-level relation with 20 virtual answers. The third \textit{Verbalizer 3} assigns a virtual answer to each third-level relation with 31 virtual answers. The "granularity" of \textit{our Verbalizer} is between \textit{Verbalizer 2} and \textit{Verbalizer 3}. We conduct experiments to compare our Verbalizer with them as well as the one using substantive answer space.

\par
We can observe that \textit{our Verbalizer} attains the best performance among all designs of an answer space. For \textit{Vervalizer 1}, each virtual answer is from a top-level relation sense that may contain some second-level relations with much different semantics and different connectives (e.g., \textit{\textbf{Tempora}l.Synchronous} and \textit{\textbf{Temporal}.Asynchronous}). Virtual answers generating by these connectives of dissimilar semantics groups cannot be distinguished very well from other virtual answers. For \textit{Vervalizer 3}, it can be attributed to deficiency of instances of some fine-grained relations. For example, \textit{Vervalizer 3} assigns a virtual answer to \textit{\textbf{Contingency}.Purpose.Arg1-as-goal} which has only three instances. Virtual answers created by few instances may not well represent the third-level relations. For \textit{Vervalizer 2}, it also encounters similar problems as \textit{Vervalizer 1} and \textit{Vervalizer 3} to some extent.

\par
As for the substantive answer space, \textit{our Verbalizer} utilizes virtual answers of appropriate granularity, which are generated by average of multiple connectives and more representative than substantive answers. It can be also observed that substantive answer space performs better than \textit{Vervalizer 1}, performs similarly as \textit{Vervalizer 2} and \textit{Vervalizer 3}. It can be concluded that although virtual answer eliminates human efforts of selecting words to construct answer space, it still needs appropriate mapping rules to attain higher performance than substantive answer space.

\begin{figure}[h!]
	\centering
	\subfloat[\small{F1-score of different design of answer space with complete knowledge distillation.}]{
		\includegraphics[scale=0.28,trim=5 10 5 5,clip]{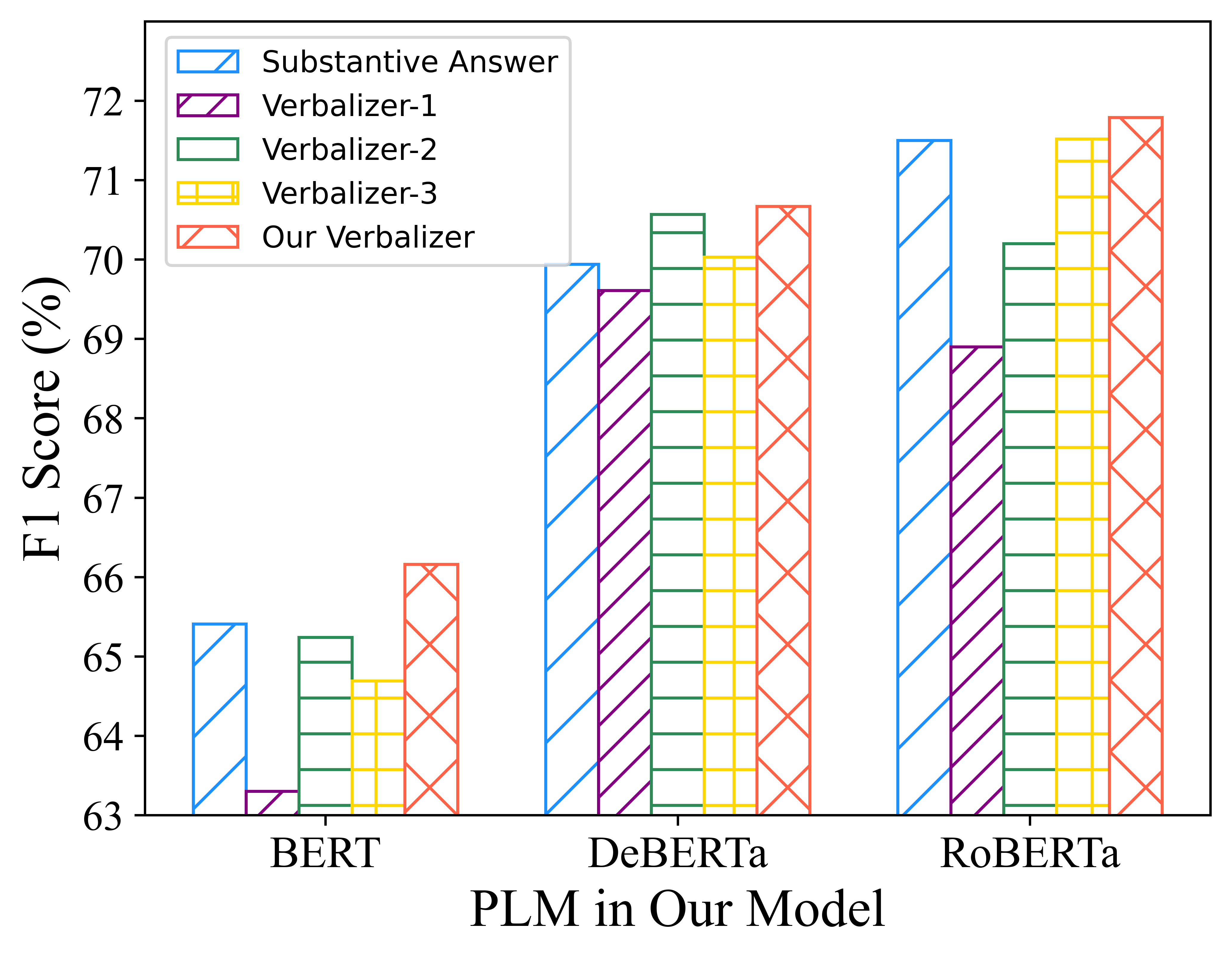}
		\label{Fig:F1 of Vervalizer with Fusion}
	}
	\quad
	\subfloat[\small{Accuracy of different design of answer space with complete knowledge distillation.}]{
		\includegraphics[scale=0.28,trim=5 10 5 5,clip]{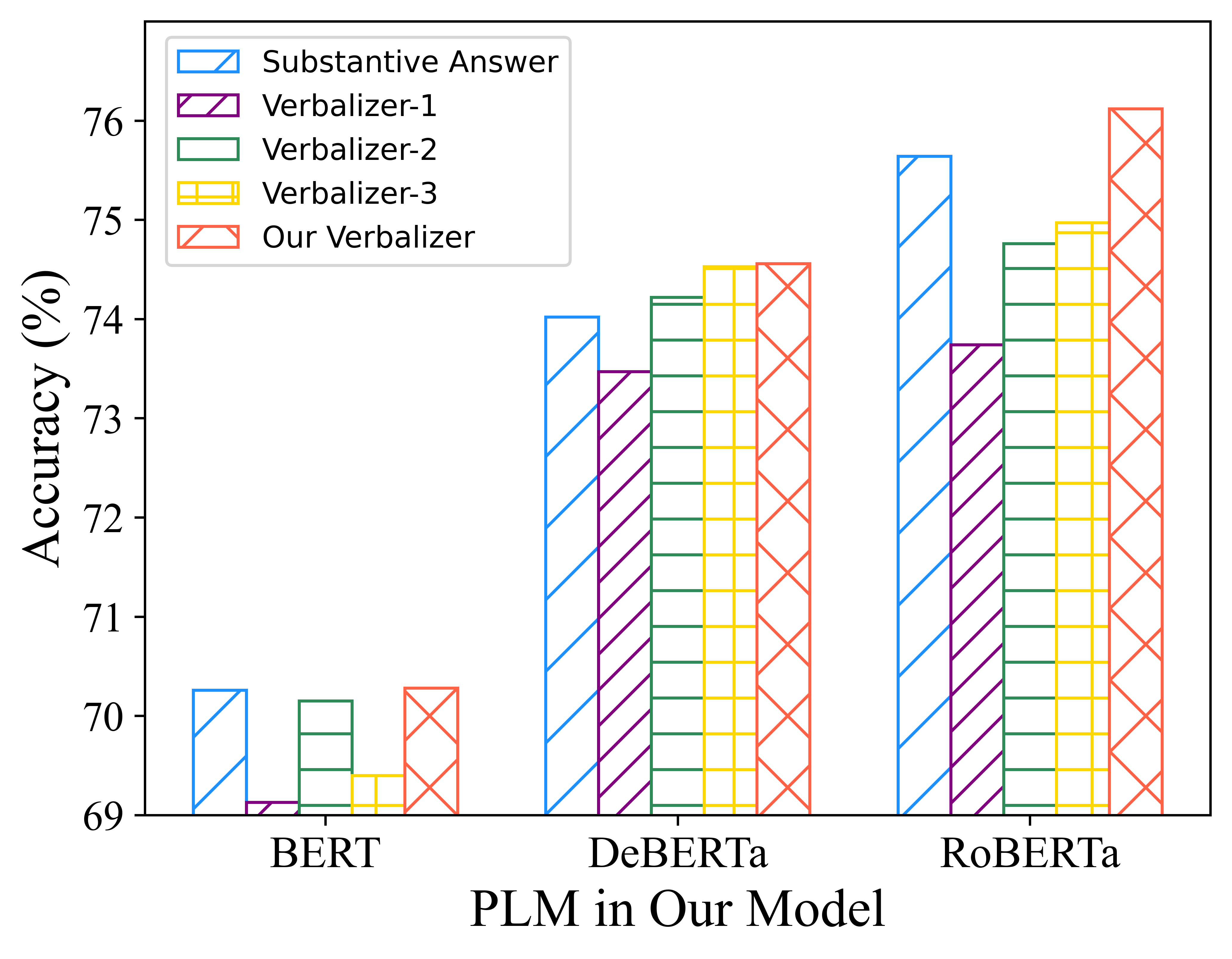}
		\label{Fig:Acc of Verbalizer with Fusion}
	}
	
	
	\subfloat[\small{F1-score of different design of answer space with response-based knowledge distillation.}]{
		\includegraphics[scale=0.28,trim=5 10 5 5,clip]{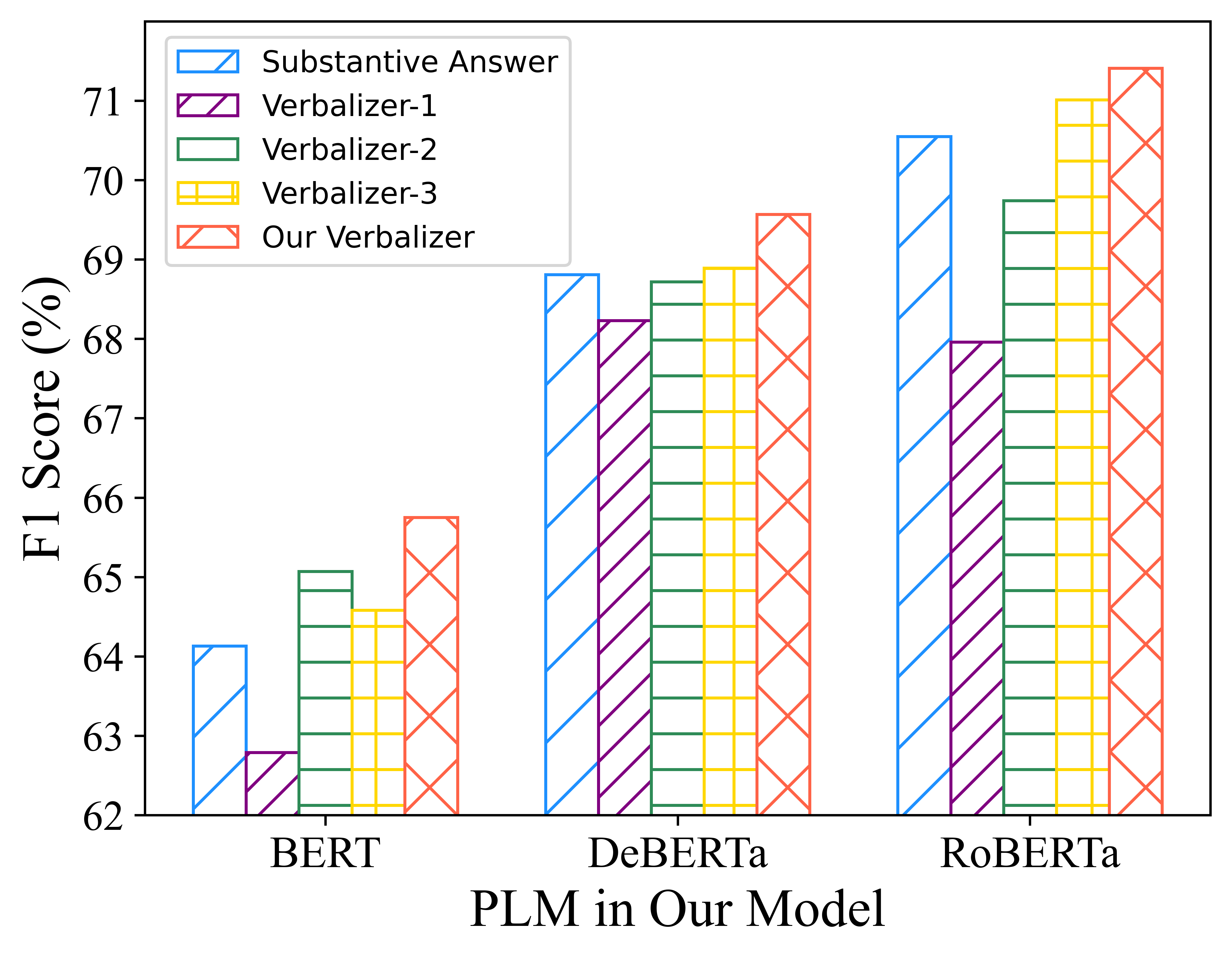}
		\label{Fig:F1 of Verbalizer with Response}
	}
	\quad
	\subfloat[\small{Accuracy of different design of answer space with response-based knowledge distillation.}]{
		\includegraphics[scale=0.28,trim=5 10 5 5,clip]{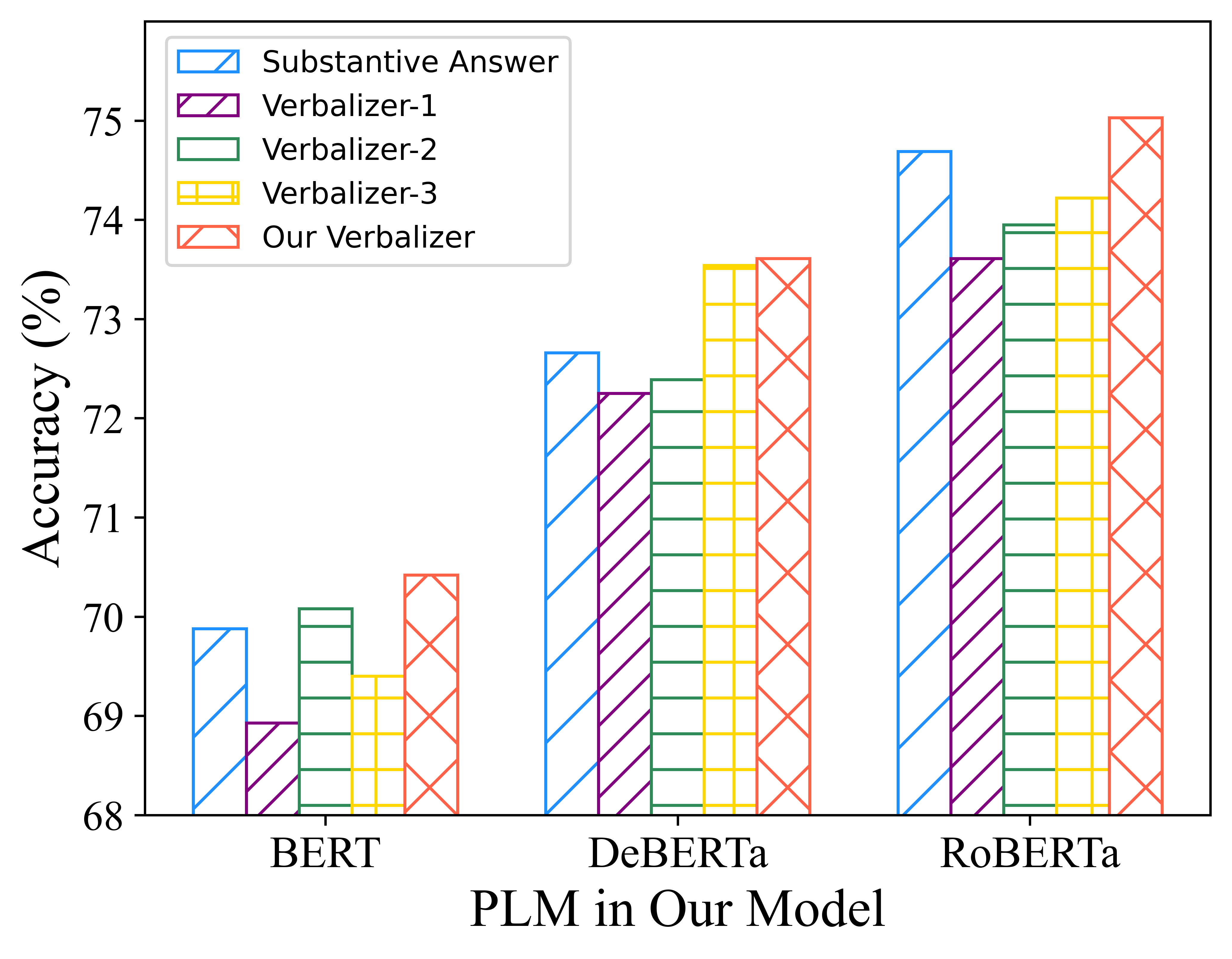}
		\label{Fig:Acc of Verbalizer with Response}
	}
	
	
	\subfloat[\small{F1-score of different design of answer space with feature-based knowledge distillation.}]{
		\includegraphics[scale=0.28,trim=5 10 5 5,clip]{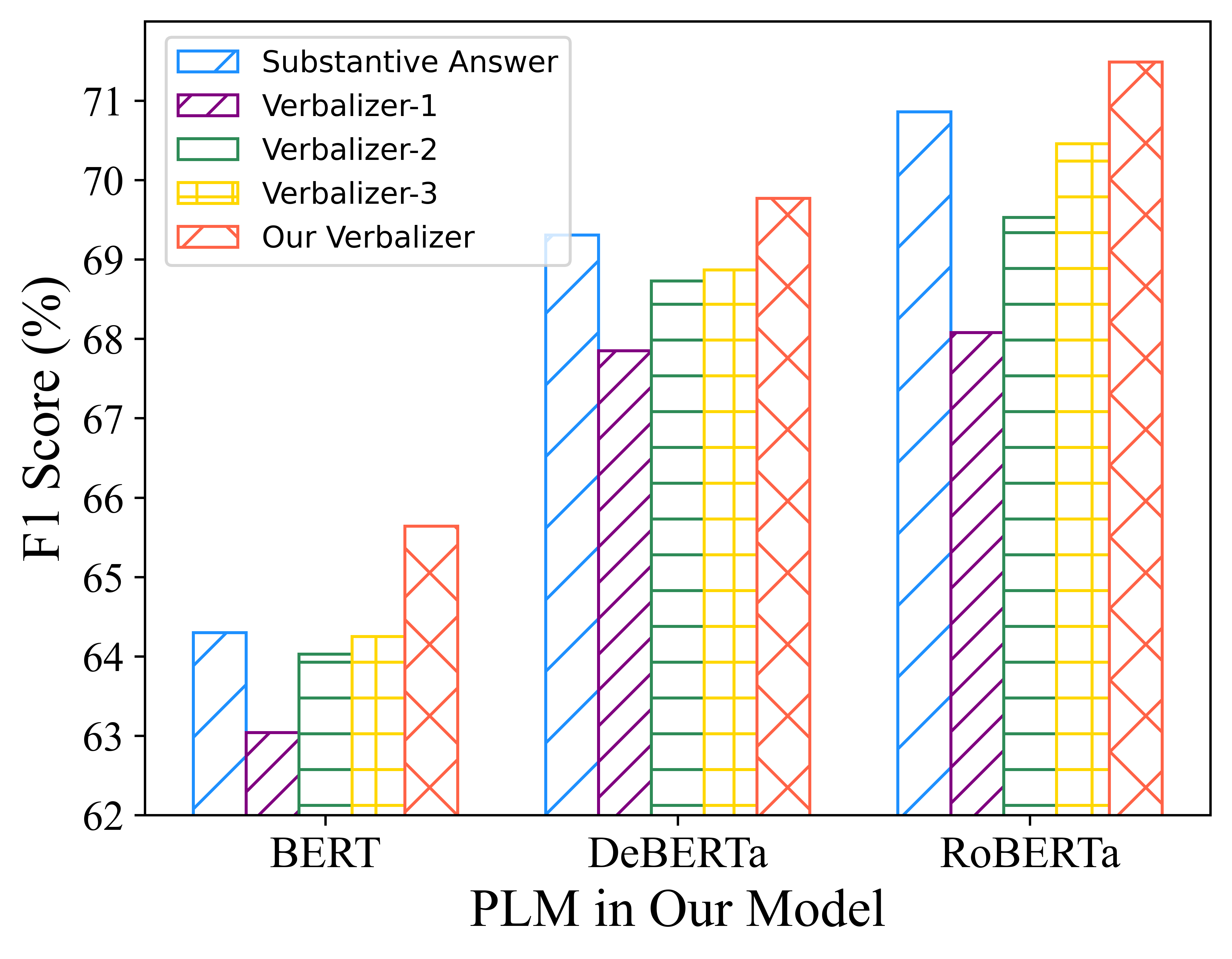}
		\label{Fig:F1 of Verbalizer with Feature}
	}
	\quad
	\subfloat[\small{Accuracy of different design of answer space with feature-based knowledge distillation.}]{
		\includegraphics[scale=0.28,trim=5 10 5 5,clip]{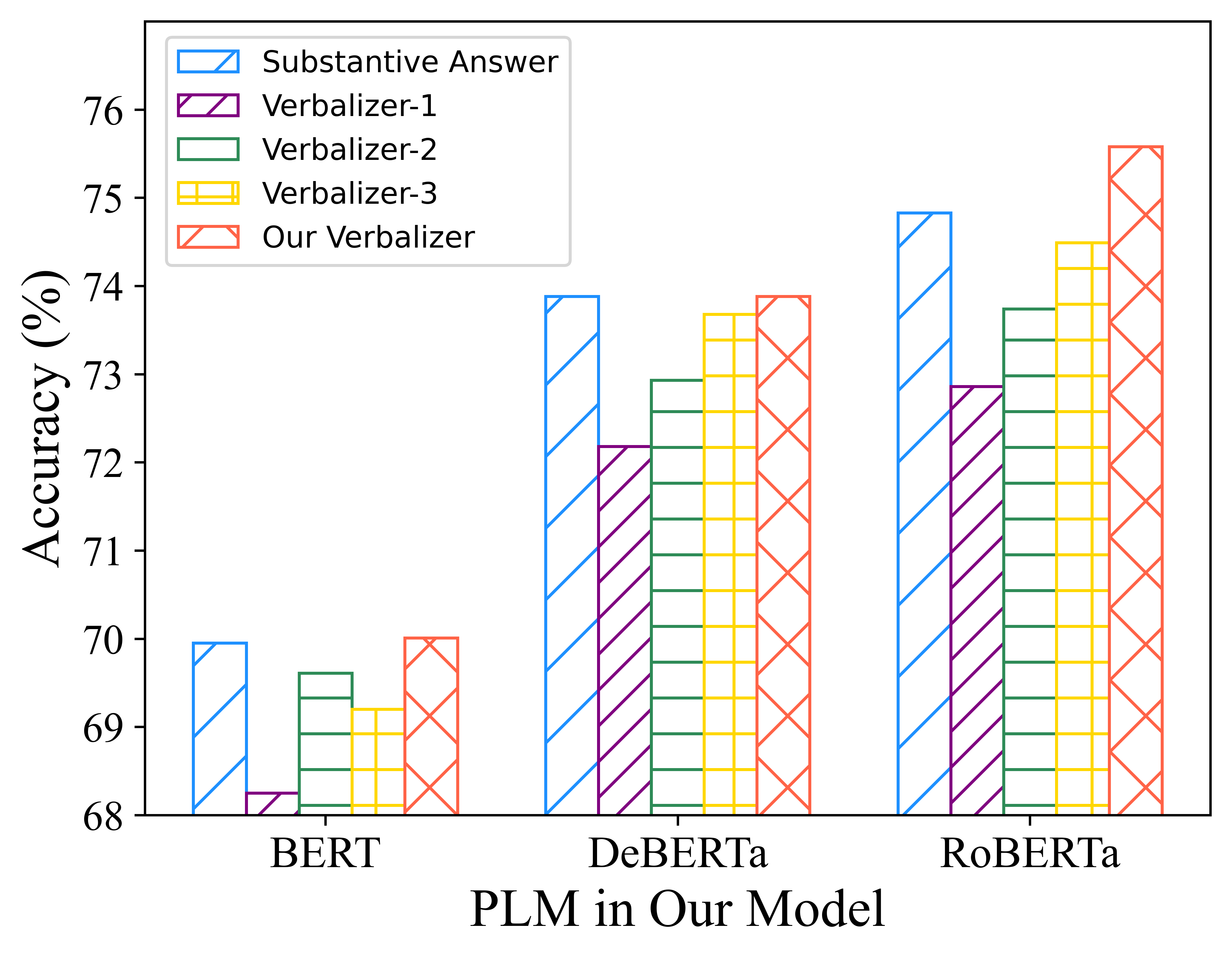}
		\label{Fig:Acc of Verbalizer with Feature}
	}
	
	\caption{The performance of different design of answer space.}
	\label{Fig:Answer Space}
\end{figure}

%
%
\section{Conclusion}\label{Sec:Conclusion}
In this paper, we have presented a novel AdaptPrompt scheme for implicit discourse relation recognition. The AdaptPrompt takes advantages of continuous prompt template to automatically search for a most suitable prompt template by gradient descent, which can better elicit prior knowledge of PLM obtained by pre-training. We also design answer-relation mapping rule and generate virtual words to construct answer space based on the mapping rule to replace traditional substantive answer space. Moreover, we utilize a teacher-student architecture to distill annotated implicit connectives' information from the teacher model to the student model to further improve the performance of our model. Extensive experiments on the latest version of PDTB 3.0 have validated that our AdaptPrompt outperforms the state-of-the-art models.

\bibliographystyle{IEEEtran}
\bibliography{ref.bib}

\section*{Appendix: The PDTB 3.0 Relations and the Answer-Relation Mapping Rule}\label{Appendix}
In this appendix, we present two tables: Table \ref{Tab:All relations of different levels in PDTB 3.0} describes relations of different levels and their corresponding relationships in PDTB 3.0 corpus; Table \ref{Tab:Virtual answer space construction rules} details answer-relation mapping rule and the answer space used in proposed model.

\begin{table*}[h!]
	\centering
	\caption{All relations of different levels in PDTB v3.0 CORPUS}
	\begin{tabular}{ |c|c|c| }
		\hline
		\textbf{Top-level} & \textbf{Second-level} & \textbf{Third-level} \\
		\textbf{Relations} & \textbf{Relations} & \textbf{Relations} \\
		\hline
		\multirow{5}{*}{\small{Comparison}} & Comparison.Similarity & Comparison.Similarity \\
		\cline{2-3}
		& Comparison.Contrast & Comparison.Contrast \\
		\cline{2-3}
		& Comparison.Concession+SpeechAct & Comparison.Concession+SpeechAct.Arg2-as-denier+SpeechAct \\
		\cline{2-3}
		& \multirow{2}{*}{Comparison.Concession} & Comparison.Concession.Arg2-as-denier \\
		& & Comparison.Concession.Arg1-as-denier \\
		\hline
		\multirow{11}{*}{\small{Contingency}} & \multirow{2}{*}{Contingency.Purpose} & Contingency.Purpose.Arg2-as-goal \\
		& & Contingency.Purpose.Arg1-as-goal \\
		\cline{2-3}
		& Contingency.Condition+SpeechAct & Contingency.Condition+SpeechAct \\
		\cline{2-3}
		& \multirow{2}{*}{Contingency.Condition} & Contingency.Condition.Arg2-as-cond \\
		& & Contingency.Condition.Arg1-as-cond \\
		\cline{2-3}
		& \multirow{2}{*}{Contingency.Cause+SpeechAct} & Contingency.Cause+SpeechAct.Result+SpeechAct  \\
		& & Contingency.Cause+SpeechAct.Reason+SpeechAct \\
		\cline{2-3}
		& \multirow{2}{*}{Contingency.Cause+Belief} & Contingency.Cause+Belief.Result+Belief \\
		&  & Contingency.Cause+Belief.Reason+Belief \\
		\cline{2-3}
		& \multirow{2}{*}{Contingency.Cause} & Contingency.Cause.Result \\
		&  & Contingency.Cause.Reason \\
		\hline
		\multirow{12}{*}{\small{Expansion}} & Expansion.Substitution & Expansion.Substitution.Arg2-as-subst \\
		\cline{2-3}
		& \multirow{2}{*}{Expansion.Manner} & Expansion.Manner.Arg2-as-manner \\
		& & Expansion.Manner.Arg1-as-manner \\
		\cline{2-3}
		& \multirow{2}{*}{Expansion.Level-of-detail} & Expansion.Level-of-detail.Arg2-as-detail \\
		& & Expansion.Level-of-detail.Arg1-as-detail \\
		\cline{2-3}
		& \multirow{2}{*}{Expansion.Instantiation} & Expansion.Instantiation.Arg2-as-instance \\
		& & Expansion.Instantiation.Arg1-as-instance \\
		\cline{2-3}
		& \multirow{2}{*}{Expansion.Exception} & Expansion.Exception.Arg2-as-excpt \\
		& & Expansion.Exception.Arg1-as-excpt \\
		\cline{2-3}
		& Expansion.Equivalence & Expansion.Equivalence \\
		\cline{2-3}
		& Expansion.Disjunction & Expansion.Disjunction \\
		\cline{2-3}
		& Expansion.Conjunction & Expansion.Conjunction \\
		\hline
		\multirow{3}{*}{\small{Temporal}} & Temporal.Synchronous & Temporal.Synchronous \\
		\cline{2-3}
		& \multirow{2}{*}{Temporal.Asynchronous} & Temporal.Asynchronous.Succession \\
		& & Temporal.Asynchronous.Precedence \\
		\hline
	\end{tabular}
	\label{Tab:All relations of different levels in PDTB 3.0}
\end{table*}

\begin{table*}[h!]
	\centering
	\caption{Details of answer-relation mapping rule and corresponding virtual answers}
	\begin{tabular}{ |c|c|c|c| }
		\hline
		\multirow{2}{*}{\textbf{Top-level Relations}}  & \multirow{2}{*}{\textbf{Third-level Relations}} & \textbf{Instance} & \textbf{Virtual} \\
		& & \textbf{Number} & \textbf{Answer} \\
		\hline
		\multirow{5}{*}{\small{Comparison}} & Comparison.Similarity & 24 & $\mathrm{[A]_1}$ \\
		\cline{3-4}
		& Comparison.Contrast & 742 & $\mathrm{[A]_2}$  \\
		\cline{3-4}
		& Comparison.Concession+SpeechAct.Arg2-as-denier+SpeechAct & 6 & \multirow{2}{*}{$\mathrm{[A]_3}$} \\
		\cline{3-3}
		& Comparison.Concession.Arg2-as-denier & 1125 & \\
		\cline{3-4}
		& Comparison.Concession.Arg1-as-denier & 40 & $\mathrm{[A]_4}$ \\
		\hline
		\multirow{11}{*}{\small{Contingency}} & Contingency.Purpose.Arg2-as-goal & 1102 & \multirow{2}{*}{$\mathrm{[A]_5}$} \\
		\cline{3-3}
		& Contingency.Purpose.Arg1-as-goal & 3 & \\
		\cline{3-4}
		& Contingency.Condition+SpeechAct & 2 & \multirow{3}{*}{$\mathrm{[A]_6}$} \\
		\cline{3-3}
		& Contingency.Condition.Arg2-as-cond & 149 & \\
		\cline{3-3}
		& Contingency.Condition.Arg1-as-cond & 3 & \\
		\cline{3-4}
		& Contingency.Cause+SpeechAct.Result+SpeechAct & 5 & \multirow{3}{*}{$\mathrm{[A]_7}$} \\
		\cline{3-3}
		& Contingency.Cause+Belief.Result+Belief & 48 & \\
		\cline{3-3}
		& Contingency.Cause.Result & 2176 & \\
		\cline{3-4}
		& Contingency.Cause+SpeechAct.Reason+SpeechAct & 9 & \multirow{3}{*}{$\mathrm{[A]_8}$} \\
		\cline{3-3}
		& Contingency.Cause+Belief.Reason+Belief & 111 & \\
		\cline{3-3}
		& Contingency.Cause.Reason & 2308 & \\
		\hline
		\multirow{12}{*}{\small{Expansion}} & Expansion.Substitution.Arg2-as-subst & 342 & $\mathrm{[A]_9}$ \\
		\cline{3-4}
		& Expansion.Manner.Arg2-as-manner & 138 & $\mathrm{[A]_{10}}$ \\
		\cline{3-4}
		& Expansion.Manner.Arg1-as-manner & 535 & $\mathrm{[A]_{11}}$ \\
		\cline{3-4}
		& Expansion.Level-of-detail.Arg2-as-detail & 2396 & $\mathrm{[A]_{12}}$ \\
		\cline{3-4}
		& Expansion.Level-of-detail.Arg1-as-detail & 205 & $\mathrm{[A]_{13}}$ \\
		\cline{3-4}
		& Expansion.Instantiation.Arg2-as-instance & 1162 & \multirow{2}{*}{$\mathrm{[A]_{14}}$} \\
		\cline{3-3}
		& Expansion.Instantiation.Arg1-as-instance & 1 & \\
		\cline{3-4}
		& Expansion.Exception.Arg2-as-excpt & 2 & \multirow{2}{*}{$\mathrm{[A]_{15}}$} \\
		\cline{3-3}
		& Expansion.Exception.Arg1-as-excpt & 2 & \\
		\cline{3-4}
		& Expansion.Equivalence & 254 & $\mathrm{[A]_{16}}$ \\
		\cline{3-4}
		& Expansion.Disjunction & 22 & $\mathrm{[A]_{17}}$ \\
		\cline{3-4}
		& Expansion.Conjunction & 3586 & $\mathrm{[A]_{18}}$ \\
		\hline
		\multirow{3}{*}{\small{Temporal}} & Temporal.Synchronous & 436 & $\mathrm{[A]_{19}}$ \\
		\cline{3-4}
		& Temporal.Asynchronous.Succession & 164 & $\mathrm{[A]_{20}}$ \\
		\cline{3-4}
		& Temporal.Asynchronous.Precedence & 847 & $\mathrm{[A]_{21}}$ \\
		\hline
	\end{tabular}
	\label{Tab:Virtual answer space construction rules}
\end{table*}



\vfill

\end{document}